\documentclass{vgtc}                          

\ifpdf
  \pdfoutput=1\relax                   
  \usepackage{graphicx}                
  \DeclareGraphicsExtensions{.pdf,.png,.jpg,.jpeg} 
\else
  \usepackage{graphicx}                
  \DeclareGraphicsExtensions{.eps}     
\fi%

\graphicspath{{figures/}{pictures/}{images/}{./}} 

\usepackage{microtype}                 
\PassOptionsToPackage{warn}{textcomp}  
\usepackage{textcomp}                  
\usepackage{mathptmx}                  
\usepackage{times}                     
\usepackage{cite}                      
\usepackage{color}
\usepackage{tabu}                      
\usepackage{booktabs}                  
\usepackage{makecell}
\usepackage{subcaption}

\usepackage{threeparttable}
\usepackage{hyperref}
\usepackage{algorithm}
\usepackage{algpseudocode}
\usepackage{amsmath}
\usepackage{amsfonts}
\DeclareMathOperator*{\argmin}{arg\,min}
\DeclareMathOperator*{\argmax}{arg\,max}
\newcommand{\Break}{\textbf{break}}
\usepackage{mathtools}
\DeclarePairedDelimiter\abs{\lvert}{\rvert}
\DeclarePairedDelimiter\norm{\lVert}{\rVert}

\usepackage{color} 
\newcommand{\hb}[1]{{\color{black}            {#1}}}

\onlineid{1260}

\vgtccategory{Research}

\title{Scale-aware Insertion of Virtual Objects in Monocular Videos}

\author{Songhai Zhang\thanks{e-mail: shz@tsinghua.edu.cn}\\
        \scriptsize Tsinghua University\\BNRist
\and Xiangli Li\thanks{e-mail:lixl19@mails.tsinghua.edu.cn}\\ %
        \scriptsize Tsinghua University %
\and Yingtian Liu\thanks{e-mail:lytlogic@gmail.com}\\ %
        \scriptsize Tsinghua University %
\and Hongbo Fu\thanks{e-mail: hongbofu@cityu.edu.hk}\\ %
        \scriptsize City University of Hong Kong }

\teaser{
 \includegraphics[height=1.2in]{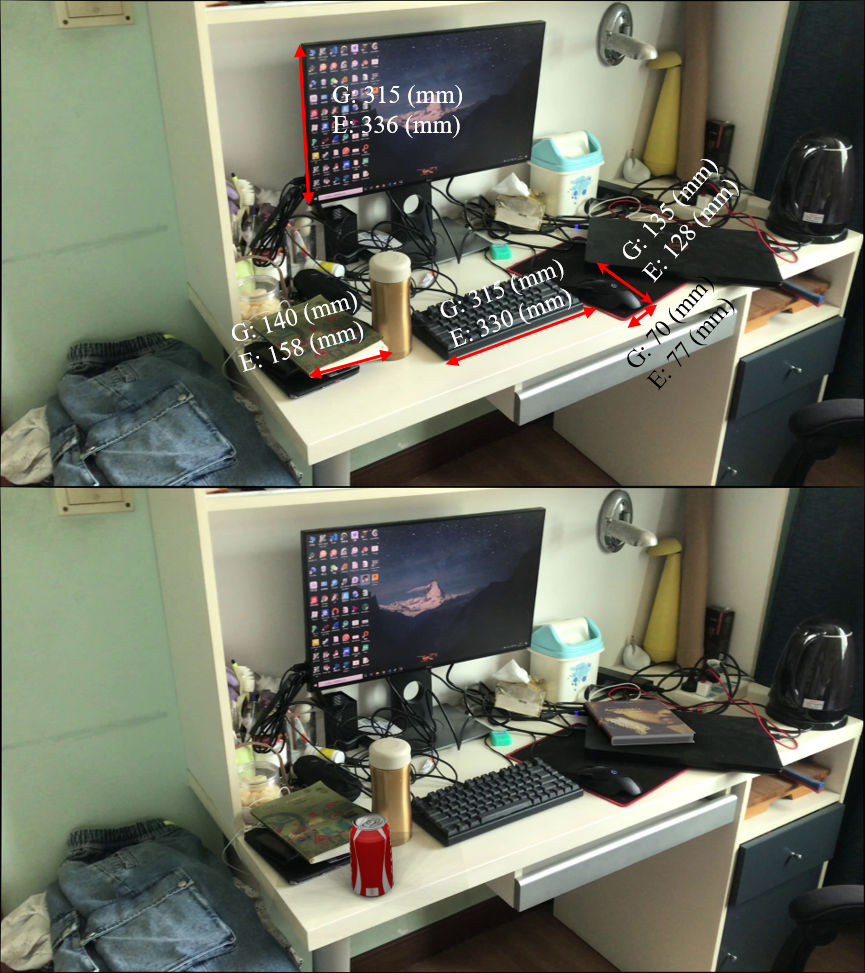}
 \includegraphics[height=1.2in]{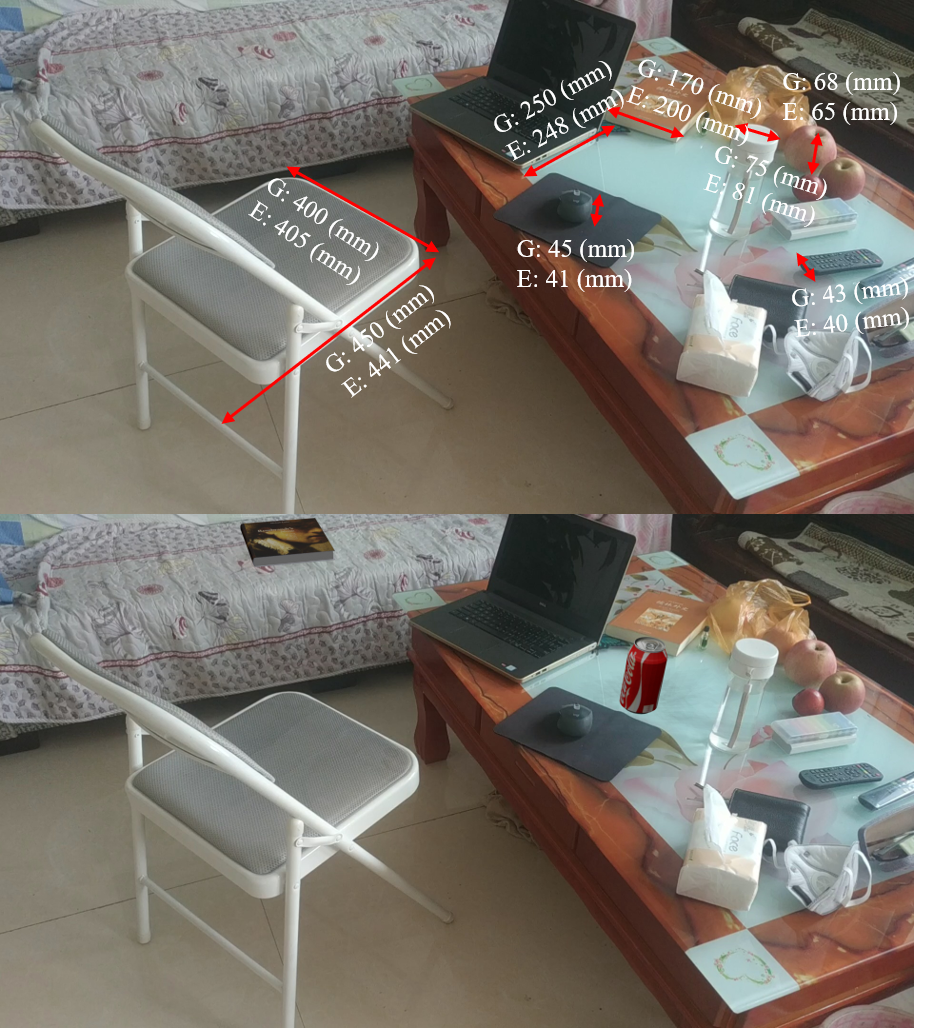}
 \includegraphics[height=1.2in]{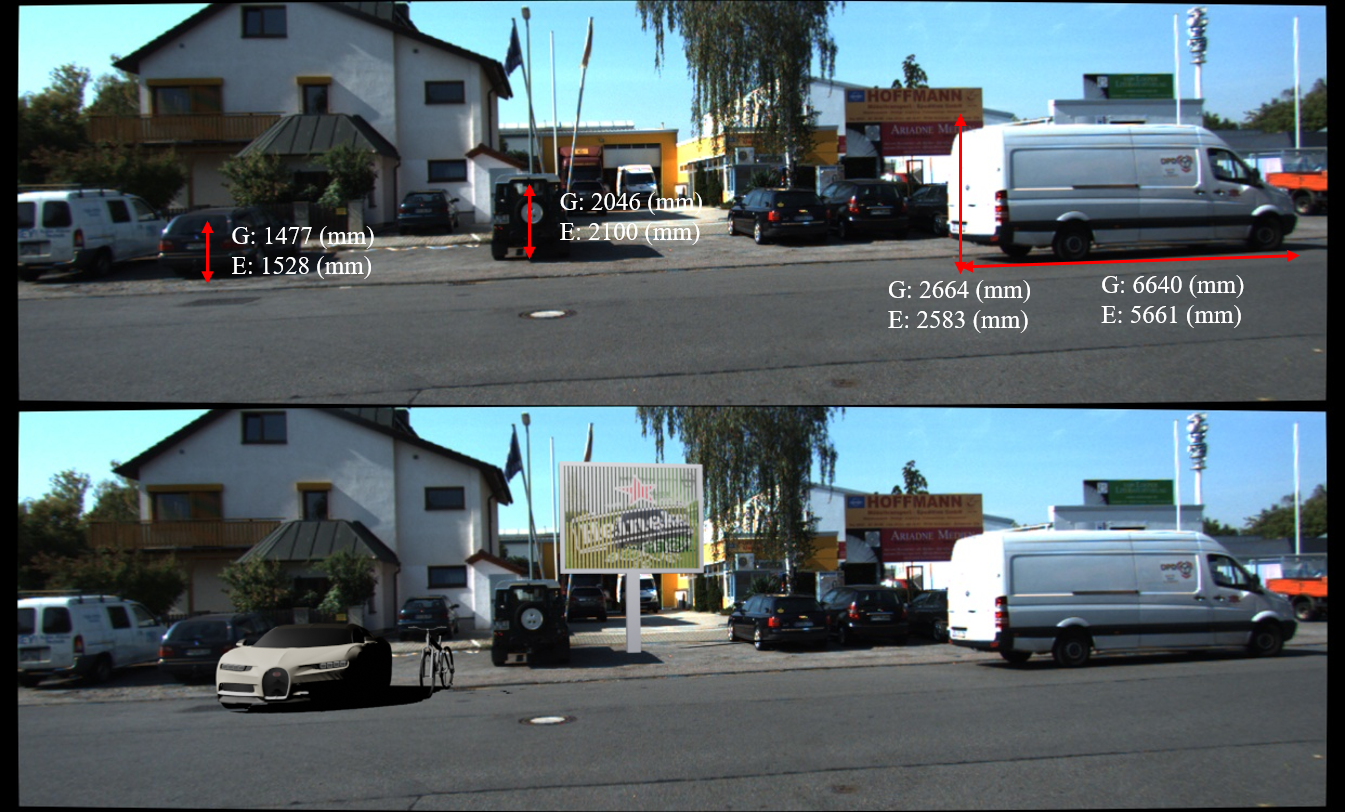}
 \includegraphics[height=1.2in]{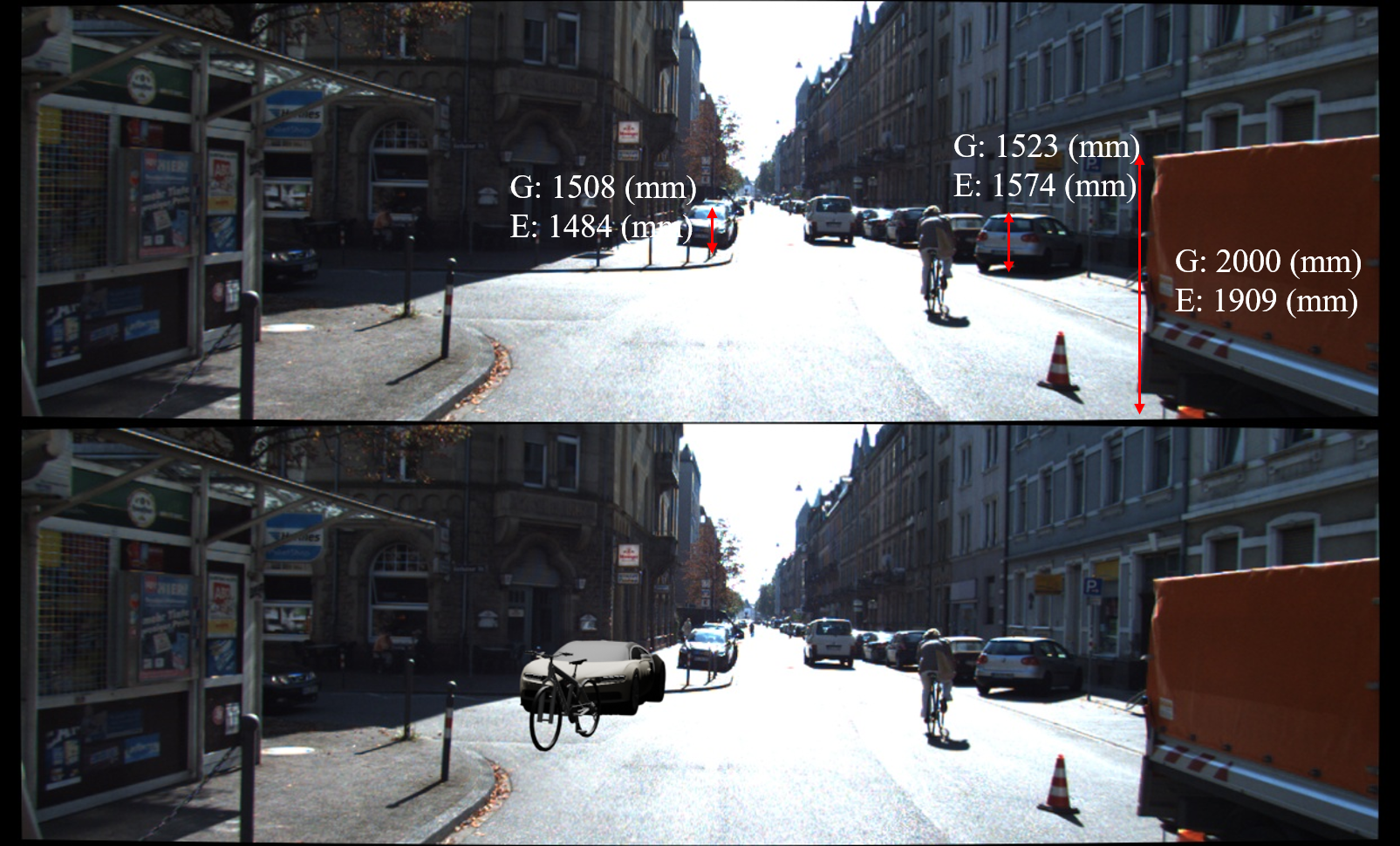}
 \caption{Scale-aware insertion of virtual objects into videos. Top: Input frames with extracted dimension and optimized size information of objects in videos (`G' for the ground-truth sizes, and `E' for the estimated sizes). The red lines indicate the extracted plausible dimensions of the objects. Bottom: Output frames with different objects inserted with proper sizes. }
 \label{fig:teaser}
}

\abstract{%
In this paper, we propose a scale-aware method for inserting virtual objects with proper sizes into monocular videos. To tackle the scale ambiguity problem of geometry recovery from monocular videos, we estimate the global scale objects in a video with a Bayesian approach incorporating the size priors of objects, where the scene objects sizes should strictly conform to the same global scale and the possibilities of global scales are maximized according to the size distribution of object categories. To do so, we propose a dataset of sizes of object categories: Metric-Tree, a hierarchical representation of sizes of more than 900 object categories with the corresponding images. To handle the incompleteness of objects recovered from videos, we propose a novel scale estimation method that extracts plausible dimensions of objects for scale optimization. Experiments have shown that our method for scale estimation performs better than the state-of-the-art methods, and has considerable validity and robustness for different video scenes. Metric-Tree has been made available at: \url{https://metric-tree.github.io}

} 

\CCScatlist{
  \CCScatTwelve{Computing methodologies}{Computer graphics}{Graphics systems and interfaces}{}
}

\begin{document}



\maketitle

\section{Introduction} 

As one of most important research problems in computer graphics and VR/AR, inserting virtual objects into real video scenes has enormous applications for both individual users and mass video owners. 
For example, automatic insertion of virtual objects into mass videos brings advertisers and content owners new opportunities by monetizing their video assets.
Seamless merging of virtual objects into videos should take into account many aspects, including scene geometry recovery\cite{10.1145/3177853,8575773}, illumination recovery\cite{Kronander2015}, rendering\cite{Kronander2015}, and an inserted object's position\cite{zhang2020} and sizing\cite{sucar:2017:probabilistic}.  
However, most of the videos capturing real-world scenes are captured by monocular cameras often without any recording camera parameters, and thus existing geometry recovering methods \cite{10.1145/3177853,8575773} often fail to recover the actual scene metrics of such videos. This is known as the scale ambiguity problem.

\begin{figure}
    \centering
    \includegraphics[width=.45\columnwidth]{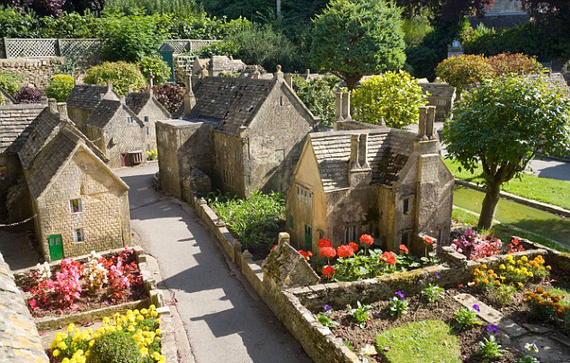}
    \includegraphics[width=.45\columnwidth]{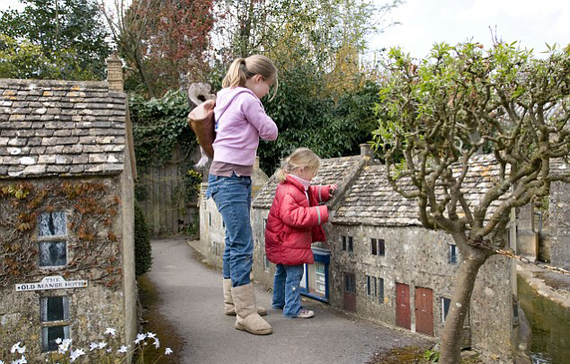}
    \caption{The miniature brings fake illusion to people on the size of a scene.}
    \label{fig:fakeillusion}
\end{figure}

We observe that the perception of an actual size of a scene in an image mainly depends on the knowledge of the sizes of objects in the scene\cite{Konkle2012A} (see Fig. \ref{fig:fakeillusion}). There are certain regular sizes of many objects in human and natural environments, such as the fixed size of A4 papers, limited size choices for beds, and a limited size range for chairs due to their use for human users. 
Such sizes following the customs of humans and size distribution statistics of natural objects, bring the background knowledge for size perception and thus make it possible to automatically estimate the actual sizes of scenes in monocular videos. 
\hb{This motivates us to design}
a two-stage system to estimate the scaling factor between a physical (or actual) size and the size of a 3D scene recovered from a monocular video. Our approach first extracts the plausible dimensions of objects from semantic segmentation of a 3D point cloud reconstructed from an input monocular video 
and then optimizes the scaling factor by incorporating the actual size distribution of object categories. In this way, the actual scene size in a monocular video can be calculated and virtual objects can be inserted with proper sizes.

Some pioneer works\cite{sucar:2017:probabilistic,9022554} have shown the effectiveness of such hints in scale estimation in scene recovery from monocular videos, and discovered at least two difficulties, making the problem non-trivial:
\begin{itemize}
    \item Objects might be partially detected or partially visible, making their size estimates inaccurate.
    \item The scale estimation decreases in accuracy due to intra-class variations in size, or even fails if no object is detected.
    \end{itemize}
We propose the following novel strategies to cope with these two issues, and make scale-aware insertion of virtual objects more automatic. 

Limited capturing views often cause incompleteness or inaccuracy of object geometry recovery by sparse or dense structure from motion (SFM) methods, so that the three dimensions of a 3D object detected in a video may not be all plausible to depict the size of this object. 
Observing the spatial features of incomplete objects, we extract plausible dimensions of objects from semantic segmentation of the point cloud. A key observation is that
all the lengths in the recovered geometry divided by their physical sizes should strictly conform to a global scaling factor. As the variance in sizes of object categories, we optimize the scaling factor by maximizing the likelihood of lengths of extracted dimensions divided by their physical sizes according to the size distributions of objects of these dimensions. 

The richness of object categories and the accuracy of their size distributions are crucial to estimate the scaling factor. The existing works take advantage of size priors by indicating the heights of several object categories, such as bottles \cite{sucar:2017:probabilistic}. In order to put it in application, we collect the physical size prior of different semantic objects from Amazon and Wikipedia, and build \emph{Metric-Tree}, a hierarchical representation of sizes of objects according to their category trees. Metric-Tree has five hierarchical levels and more than 900 object categories as leaf nodes, covering the furniture, car, electric appliance, person, and so on. For each category, we build a Gaussian Mixture Model (GMM) model on 3D sizes of the associated objects according to the collected size data. 

The results of different interior and outdoor scenes show that although every object category contains variance in size, the scale estimation error rapidly decreases along with the number of objects involved in scale optimization. The scale estimation error can be reached less than 8\% in average in the Kitti dataset for outdoor videos and our captured indoor-scene videos in daily lives.

In summary, our work has two contributions:
\begin{itemize}
    \item We propose a large-scale dataset of sizes of object categories: Metric-Tree, a hierarchical representation of sizes of more than 900 object categories as well as the corresponding images, connecting the appearance and the size of each object category.
    \item We propose a novel scale estimation method that extracts plausible dimensions of objects for scaling factor optimization to alleviate the incompleteness of geometry recovery from monocular videos.  
\end{itemize}

\section{Related work}
\subsection{Object insertion in videos}
Synthesizing realistic images or videos by merging virtual objects into real world scenes seamlessly is one of the longstanding goals of computer graphics and one of the main applications in AR/VR. Although there is still no completely automatic solution  
as far as we know, different aspects of relative research have been taken out to make this process more intelligent and thus automatic. 
The underlying geometry of scenes in videos can be recovered by structure from motion (SFM) and visual simultaneous localization and mapping (vSLAM) techniques\cite{10.1145/3177853,8575773}. 
Capturing, estimating and rendering with scene illumination are summarized by \cite{Kronander2015}. Context-based recommendation for object insertion in visual scenes is a relatively new topic, 
and there have been some pioneer works on recommendation in image \cite{zhang2020} by modeling the joint probability distribution of object categories, and object recommendation systems by neural networks \cite{wang2019planit}. For automatic insertion of virtual objects into monocular videos, the size of inserted objects is a critical factor to influence the photo-realistic effects of the resulting videos. However, the scale ambiguity problem for monocular videos has been largely unexplored and is the focus of our work. 

\subsection{Scale estimation}

Due to the classical problem of scale ambiguity in the 3D reconstruction from monocular videos, at least one scale-related piece of knowledge needs to be introduced to recover the actual size of the overall scene. 
Some methods combine sensors, such as inertial measurement units (IMUs) \cite{nutzi:2011:fusion,martinelli:2011:vision,grabe:2013:comparison,ham2014hand,garro2016fast}and LiDAR \cite{brenneke:2003:using,zhang:2015:visual} into SLAM systems to estimate the unknown scaling factor.
Other methods incorporate camera setup information as priors into SLAM systems, such as camera height in a car based on the ground plane assumption \cite{song:2014:robust,song:2016:high,zhou:2019:ground,grater:2015:robust,zhou:2016:reliable} and the information of the camera's offset relative to a vehicle's axis of rotation when the vehicle turns \cite{scaramuzza:2009:absolute}.
Those methods with additional sensors or camera setup information achieve impressive results,whereas they do not meet our need for addressing the scale estimation problem for monocular videos in absence of the camera parameters.

In the autonomous driving area, a large number of visual odometry systems incorporate semantic information by object tracking or instance segmentation to address the scale drift problem \cite{botterill:2013:correcting,hilsenbeck:2012:scale,sucar2018bayesian,lianos2018vso}, where the basic idea is to find semantic-level feature correspondences among key frames and combine them with feature matching into bundle adjustment. Their goal is to alleviate the scale drift same as the loop closure, and our method is to estimate the actual size of a scene captured in a monocular video.

Ku et al. \cite{Jason2019Monocular} proposed a 3D object detection method from an image, and derive 3D bounding boxes of three object categories including car, pedestrian and cyclist, in actual sizes. To achieve this, they take advantage of the LiDAR data in training to learn the shape and scale information. The strong prior knowledge of point clouds of scenes provides pretty good estimation on object sizes but also limits the application of their approach to a wider range of scenes. Our method makes use of a new dataset of sizes of object categories  and their corresponding images for instance segmentation and scaling factor estimation. Our dataset is much easier for expansion than point clouds to cope with many more new types of objects. 

Sucar et al. \cite{sucar:2017:probabilistic} present a pioneer work on scale estimation from monocular videos, \hb{and their approach is the closest to ours}. They use the YOLO v2 network for the object recognition task and project an object's bounding box in an image frame into a 3D scene to calculate the object's height and thus estimate the scaling factor based on the assumed height distribution. They experimentally demonstrate the feasibility of their method under ideal conditions, but their performance on real object size distributions has not been evaluated. One shortcoming with their approach is its prescription for the vertical orientation of the scene. 
We propose a more advanced method for plausible dimension extraction of objects and incorporate a size dataset of object categories for scaling factor optimization, thus achieving more accurate results (see Section \ref{sec:compare}).

\subsection{Datasets of object sizes}

Some existing works have collected size data for sizing the 3D shapes in a 3D collection. 
Shao at el. \cite{shao2017cross} proposed a method for transferring physical scale attributes between web pages and 3D Shapes. They leverage the text and image information to connect web pages and shapes. The former is used for matching web page text with object text to build a direct connection. The latter relies on visual similarity to construct a joint embedding space between images and shapes. Finally, scale attributes can be transferred between the closest pairs in the embedding space. 
Savva at el. \cite{savva2014being} proposed a probabilistic graphical model for sizing the collections of 3D shapes. They also collect 3,099 furniture items in 55 categories and transfer the sizes to other 3D shapes by maximizing the probability distribution with size priors of object categories and the scale consistency of co-occurrences of objects in 3D scenes in virtual scenes. Savva at el. \cite{savva2015semantically} further connected more physical attributes of objects with 3D shapes including weight, static support and attachment surfaces. We focus on the sizes and appearances of object categories in the image space, and collect a much larger dataset involving more than 900 categories with sufficient samples in each category to support  scene size understanding. Some professional websites, such as dimensions.guide\footnote{https://www.dimensions.guide} also provide sizes of everyday objects and spaces that make up the world. These websites mainly serve industrial design, and have only a few typical samples for each category.

\begin{figure}
    \centering
    \includegraphics[width=\columnwidth]{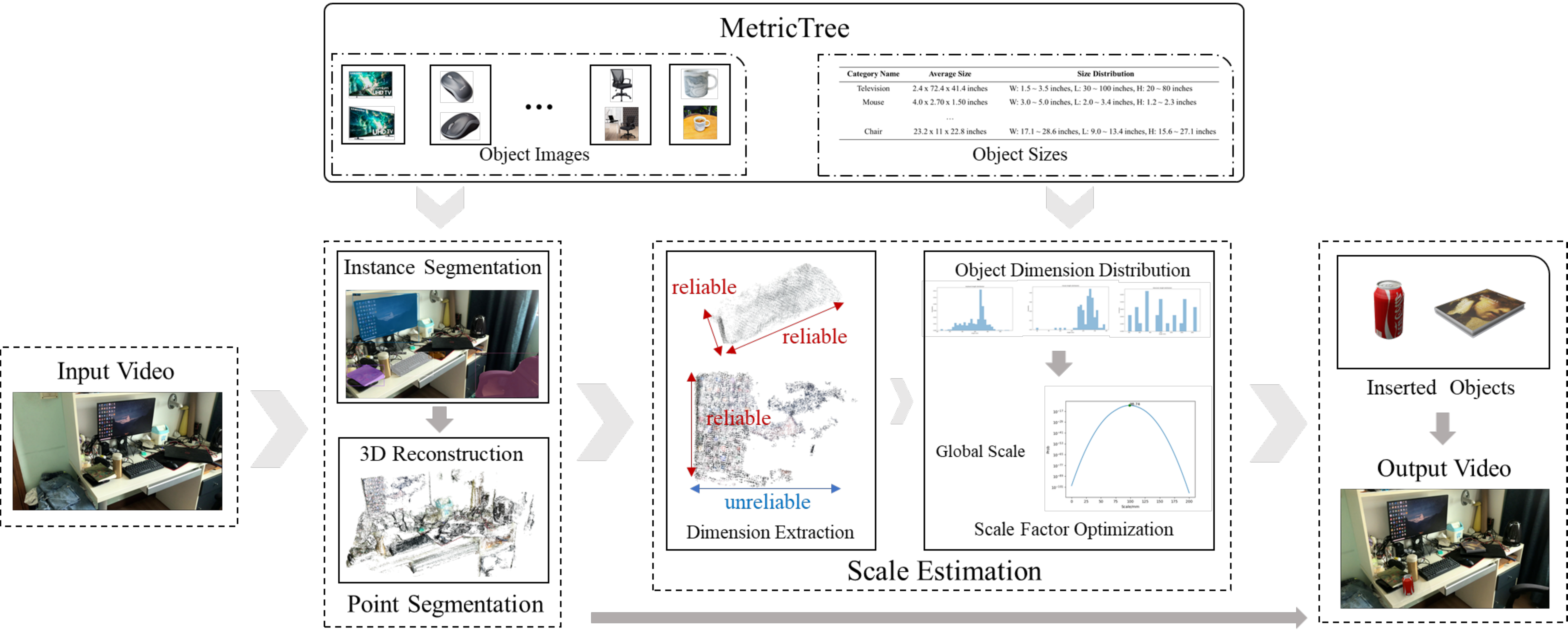}
    \caption{The pipeline of our method.}
    \label{fig:pipeline}
\end{figure}

\section{Overview}
Our system for scale-aware virtual object insertion into monocular videos follows the general pipeline of virtual object insertion, as shown in Fig. \ref{fig:pipeline}. It begins with reconstructing a 3D scene represented as a point cloud from an input monocular video \cite{opensfm}. We then perform instance segmentation of the point cloud by fusing per-frame segmentation.
The key step is scale estimation between the point cloud and the actual scene (Section \ref{sec:ScaleEstimation}). In this step, due to the incompetence of objects in the point cloud, we extract plausible dimensions of objects and optimize the scaling factor by incorporating with the priors of size distributions of the corresponding object categories. We assume the actual sizes of virtual objects are known, so that they can be inserted with proper sizes by multiplying the estimated scaling factor. Our proposed dataset Metric-Tree (Section \ref{sec:Metric-Tree}) provides not only the strong priors of size distributions to scaling factor optimization, but also provides image samples for semantic segmentation.

\begin{figure}
    \centering
    \includegraphics[width=.8\columnwidth]{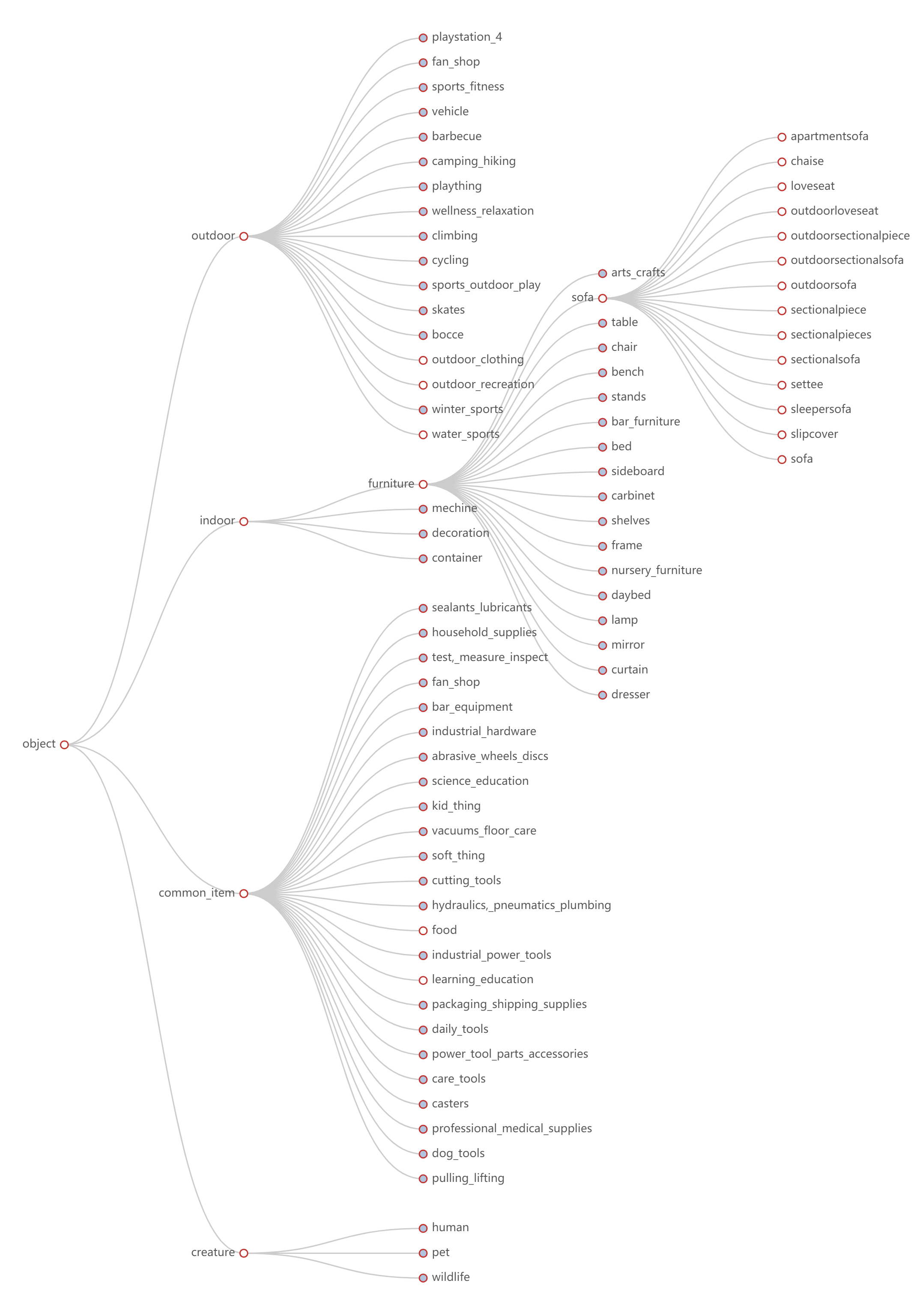}
    \caption{The structure of Metric Tree. The hollow dots are expanded nodes and the solid dots are folded nodes.}
    \label{fig:metrictree}
\end{figure}

\section{Metric-Tree: A Repository of Object Sizes}
\label{sec:Metric-Tree}

As mentioned above, the richness of object categories and the accuracy of their size distributions are crucial to estimate the scaling factor. However there is no open source repository of object dimensions, and thus we have to collect abundant sizes of objects for constructing  distributions of sizes. 
Our approach is based on three key facts and assumptions: 
\begin{itemize}
    \item In the real world, the dimensions of most objects are fixed, or with a small range of changes in some dimensions. 
    \item The dimensions of objects' minimum bounding boxes are consistent and very close to physical sizes.
    \item The sizes of objects in a real scene are interrelated, and the sizes of objects in the same scene should be consistent. 
\end{itemize}


\textbf{Webpage Data.} 
Like \cite{shao2017cross}, we collect the sizes of objects with their images and texts through the Internet. We crawl the Amazon websites to extract the physical sizes of  object categories (e.g.,  "table", "chair", etc.). 
Besides, for the other objects that do not appear in Amazon (like "car", "hydrant", "person"), we crawl from Wikipedia, Crate\&Barrel, car websites, etc. for gathering their sizes and corresponding images. 
We finally get about 10,000 raw categories, among which there are about 80,000 raw items with category annotation, and images (from the source websites we crawled, and most of them are with white background). After removing categories without any size information and merging the similar categories, we build a dataset of object sizes with a five-level tree structure based on WordNet \cite{miller1995wordnet} with about 900 categories in the leaf nodes.

\begin{figure}
    \centering
    \includegraphics[width=\columnwidth]{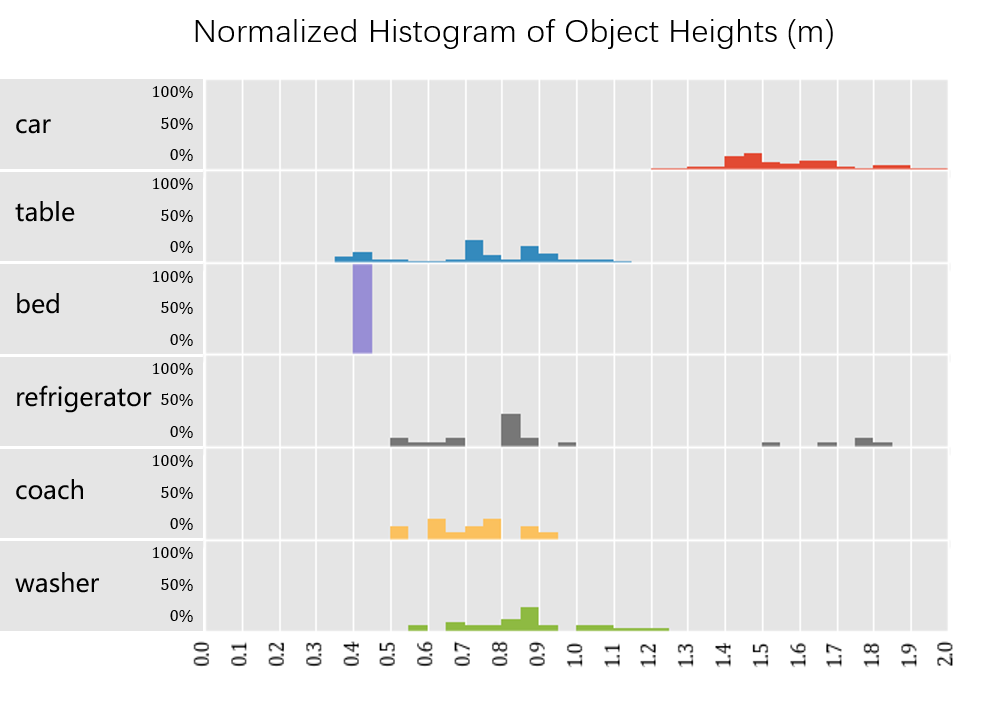}
    \caption{Size distribution of severeal typical object categories.}
    \label{fig:sizedistribution}
\end{figure}

\textbf{Metric-Tree.} As shown in Fig. \ref{fig:metrictree} (which shows the part of Metric-Tree, due to space limitations), Metric-Tree is a tree structure, with each node being an object category. Each node has two components: a set of dimension data and a dataset of  images of the corresponding object category. 
We organize the size data based on WordNet by generating a tree structure and attaching the dimensions and corresponding images to leaf nodes. For categories not included in WordNet, we use a multi-person proofreading method to insert them into the data structure.
For inner nodes, the dimension data set and image data set are all defined as the aggregate of all their children. For each category, we build a Gaussian Mixture Model (GMM) for 3D sizes of the objects as a size distribution according to its dimension data set. 

We also use BASNet \cite{Xuebin2019BASNet} to perform foregound segmentation \hb{to extract foreground objects.}

As the state-of-the-art fine-grained classification methods still do not work well, 
we retrain Mask R-CNN\cite{he:2017:mask, wu:2019:detectron2} as our instance segmentation network.
We select 43 object categories of segmentation according to the following rules:
\begin{itemize}
    \item The divergence of size distribution of object categories is low. Fig. \ref{fig:sizedistribution} illustrates the size distributions on height of several typical categories with different divergence. 
    \item The number of samples in the dimension data set is enough to depict the size distribution.
    \item There are additional training data in other famous datasets.
\end{itemize}
The traing set of the images is the combination of image datasets of corresponding object categories and correponding subsets with the label of object categories in the COCO dataset\cite{lin:2014:coco}, ADE20K\cite{zhou:2016:ade20k, zhou:2017:scene} and Open Images\cite{kuznetsova:2018:openimages}.

\section{Scale estimation with plausible dimensions}
\label{sec:ScaleEstimation}

\subsection{Instance segmentation on point clouds}

\begin{figure}
    \centering
    \includegraphics[width=.9\columnwidth]{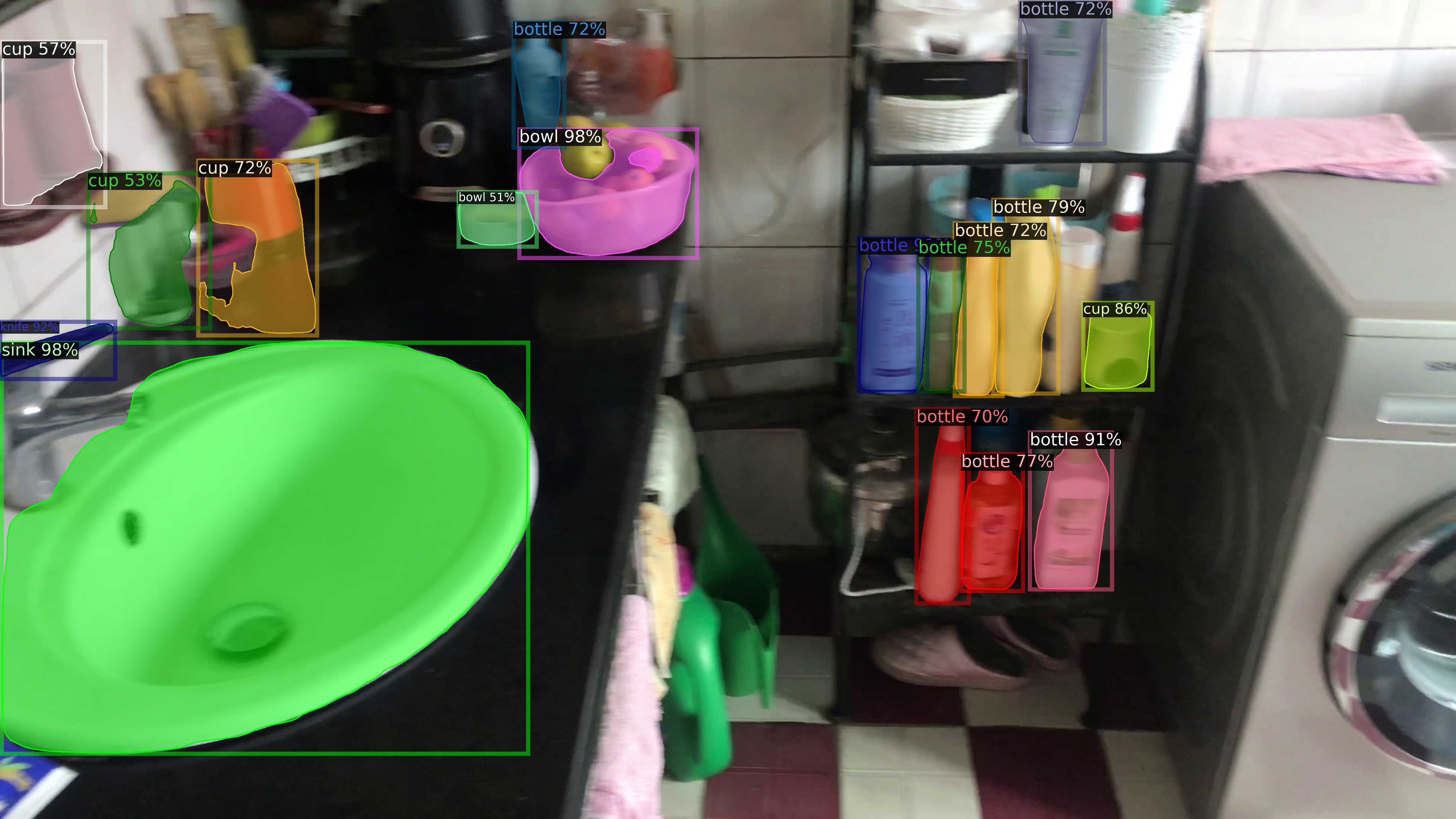}
    \caption{Instance segmentation results from Mask R-CNN.}
    \label{fig:mask-result}
\end{figure}

We use the open source OpenSfM \cite{opensfm} system integrated with the PatchMatch \cite{shen:2013:accurate} to reconstruct 3D scenes, and use Mask R-CNN\cite{he:2017:mask, wu:2019:detectron2} retrained by our collected image data(Section \ref{sec:Metric-Tree}) to perform instance segmentation(Fig. \ref{fig:mask-result}).
Although some existing methods provide instance-level point cloud segmentation, such as MaskFusion\cite{runz2018maskfusion}, given that it uses RGB-D inputs, we cannot ensure its validity with RGB inputs, so we clarify our approach below.

After 3D reconstruction and instance segmentation on frames $\{F_i\}_{i=1}^N$, for frame $F_i$, we get the point cloud $S_i$, the camera pose $C_i$ and the segmentation results $\{O_{ij}\}_{j=1}^{M_i}$, where $M_i$ is the number of recognized objects and $O_{ij}$ includes the class $c_{ij}$ of the $j$-th object $o_{ij}$ and the pixel-level mask $m_{ij}$(see Fig \ref{fig:mask-result} for example).
We map the 2D instance segmentation results to the point cloud. Each point is labeled according to the 2D instance where its projection is located. Given the point cloud $S_i$ of $i$-th frame and segmentation results $\{O_{ij}\}_{j=1}^{M_i}$, the reconstructed point cloud $s_{ij}$ of object $o_{ij}$ in this frame is
\begin{equation}
    s_{ij} = \{\ p\ |\ p\ \in S_i\ \wedge\ \hat p \in m_{ij}\},
\end{equation}
where $\hat p$ denotes the projection of $p$ in frame $i$.

The next step is to merge point clouds of the same object that are split in different frames to get the complete point cloud of that object. For a real object $A$ in the scene, let $O_{{i_1}{j_1}},\dots, O_{{i_{N_A}}{j_{N_A}}}$ denote the recognition of $A$ in different frames,  where $N_A$ is the times $A$ appears. Then the merged point cloud of $A$ is 
\begin{equation}
    S_A = \cup_{k=1}^{N_A}\ s_{{i_k}{j_k}}. 
\end{equation}

To solve the problem of correspondence of instances of the same object in multiple frames, we propose an incremental point-cloud merging method that recovers all real objects in the scene by merging point clouds frame by frame. In the following discussion, without loss of generality, we consider only the case where objects are of the same class, otherwise we can split the point clouds by classes first, since point clouds with different classes obviously do not correspond to the same objects.

Let $U = \{u_k\}_{k=1}^K$ denote the intermediate point cloud during merging process and $K$ denote the number of objects we have obtained. Initially, $U$ is empty. If the previous $g$ frames have been merged, then we consider the relationship between the current $U$ and the reconstructed point cloud $S_{g+1}$ of the $g+1$ frame: some of the point clouds in $S_{g+1}$ correspond to objects already in $U$ and some of them are new objects (the subscript $g+1$ is omitted following for brevity). Here we define the distance between point cloud $A=\{a_i\}_{i=1}^M, B=\{b_i\}_{i=1}^N$ (without loss of generality, let $M \le N$) as
\begin{equation}
    D(A, B) = \frac 1 M \sum_{i=1}^{M} \min_{\forall b_j \in B} \norm{a_ib_j}. 
\end{equation}
\hb{where} 
$M = |S|$ denotes the number of instances recognized in frame $g+1$. We calculate the distances of $u_k \in U$ and $s_j \in S$ based on above distance definition. Then we greedily look for the point cloud pair $(u_k, s_j)$ with the smallest distance and mark them as the same object until the distance $D(u_k, s_j)$ exceeds a predefined threshold or one of $U$ or $S$ has completed the match. After repeating this process frame by frame, we complete the point cloud merging(see \ref{merge} for pseudocode). The merged point cloud may have some noise resulting from incorrect feature point reconstructions or inaccurate merge, so we can obtain the main part of each object after further de-outlier operations. Here KNN\cite{cover:1967:nearest} and the isolated forest method\cite{liu:2008:isolation} are used.

\begin{figure}
    \centering
    \includegraphics[width=0.9\columnwidth]{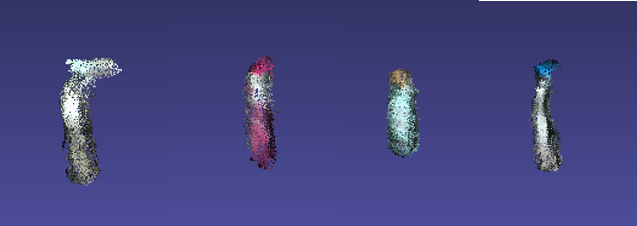}
    \caption{4 bottles are recovered incompletely due to blocking in the scene shown in Fig \ref{fig:mask-result}. Dimension extraction is required to obtain plausible dimensions. }
    \label{fig:instance-segmentation}
\end{figure}

\begin{algorithm}[htb]
    \caption{Incremental point cloud merging}
    \label{merge}
    \begin{algorithmic}[1]
        \Require
        point clouds from all frames $S = \{s_i\}_{i=1}^N$
        \Ensure
        object point clouds $U = \{u_k\}$
        \State $U \gets$ \textit{empty list}
        \For{$s_i \in S$}
        \State $U \gets$ \Call{MergePointCloud}{$U, s_i$}
        \EndFor
        \State \Return $U$
        \State
        \Function{MergePointCloud}{$U, s$}
        \State $m \gets |U|$,\ $ n \gets |s|$
        \State $\hat U \gets U$,\ $\hat s \gets s$
        \State $U \gets$ \textit{empty list}
        \While{$\hat U\ \textit{not empty} \ \textbf{and} \ \hat s\ \textit{not empty}$}
        \State $u_i, s_j \gets \argmin_{\forall u_i\ \in\ \hat U,\ s_j\ \in\ \hat s} D(u_i, s_j)$
        \If{$D(u_i, s_j) <\ \textbf{threshold}$}
        \State \Break
        \EndIf
        \State $\hat U = \hat U \setminus u_i,\ \hat s = \hat s \setminus s_j$
        \State $U = U \cup \{(u_i \cup s_j)\}$
        \EndWhile
        \State $U = U \cup \hat s$
        \State \Return $U$
        \EndFunction
    \end{algorithmic}
\end{algorithm}

\subsection{Dimension extraction for 3D objects}
\label{sec:DimExtraction}

Since in practice we mostly use the length, width and height of an object, also known as dimensions, to describe its size. To determine the orientations of the object, We first use the camera pose to determine the orientation of the bottom. In order to reduce the uncertainty of the orientation estimation, we make the following assumptions about the camera and the scene: the roll of the camera is zero, and the object is placed on a flat, horizontal surface.
Based on these assumptions, we can constrain the direction of the bottom surface of the object. Let $\{\vec {r_i}\}_{i=1}^N$ denote the $x$ axis of the camera in all frames, as well as the right directions. Then the normal vector $\vec n$ of the desired horizontal plane is the solution of this least-squares minimization problem by denoting 
$R = [r_1 r_2 \dots r_N]^T$:
\begin{equation}
    \vec n \ = \ \argmin_{\vec x\ \in \ \mathbb{R}^3,\ \norm{\vec x} = 1}\ \norm{\ R\vec x\ }. 
\end{equation} 
It is easy to know that $\vec n$ is the unit eigenvector corresponding to the minimum eigenvalue of $R^TR$.
After the bottom of the object is identified, the problem is reduced to a 2D point cloud dimension calculation, which can be solved by minimum bounding box method.

We have found that the dimension extraction of objects is susceptible to the results of reconstruction and segmentation, with the corresponding dimension of an object being inaccurate when the local point cloud is relatively sparse and incorrectly segmentation occurs. We therefore introduce dimension confidence (also called reliability) to measure the results and to provide guidance for subsequent scale optimization. A reliable reconstructed point cloud should have a similar density in all regions, so we estimate the reliability of the dimension computation based on this assumption.

\begin{figure}
    \centering
    \includegraphics[width=\columnwidth]{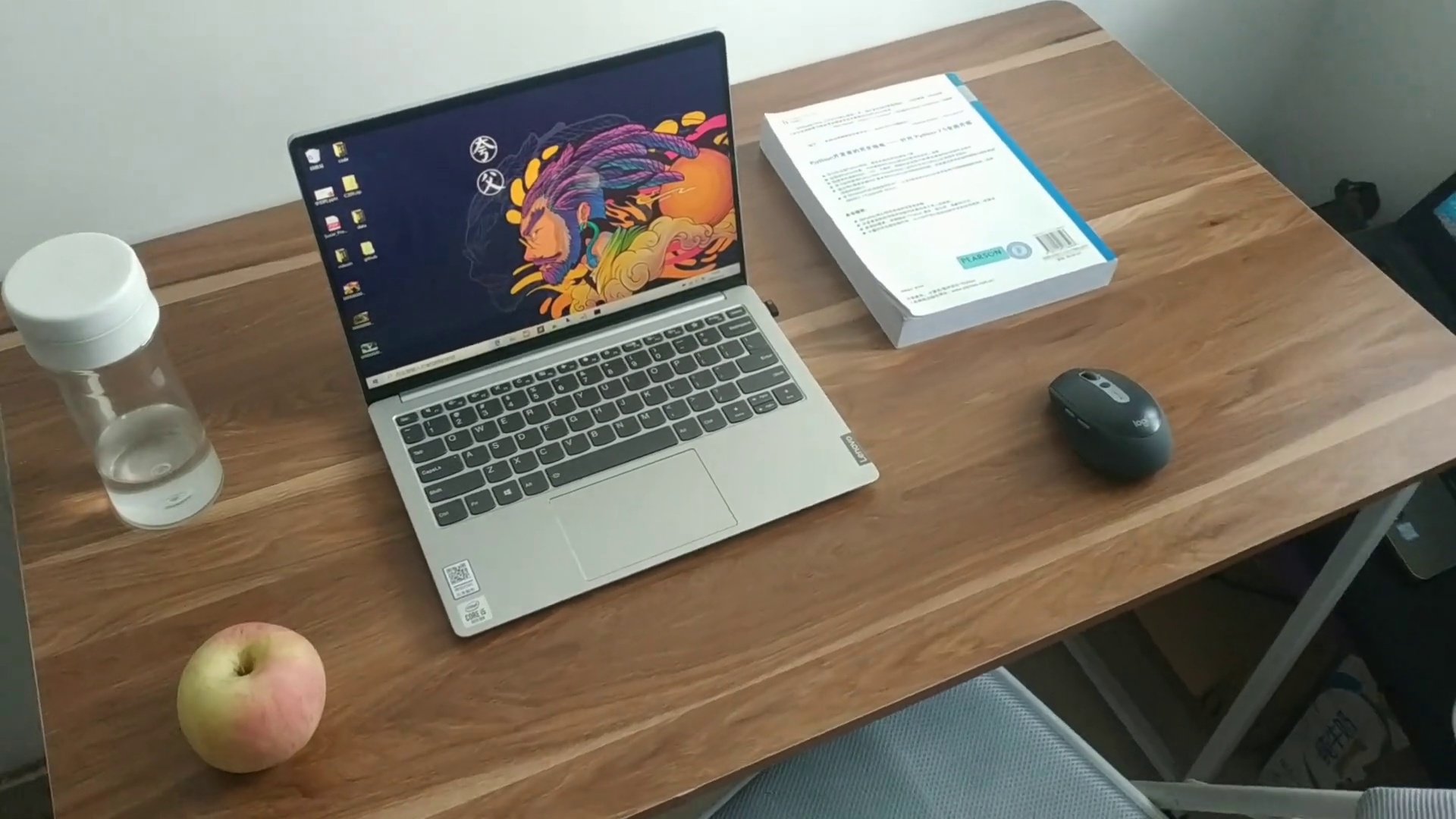}
    \includegraphics[width=\columnwidth]{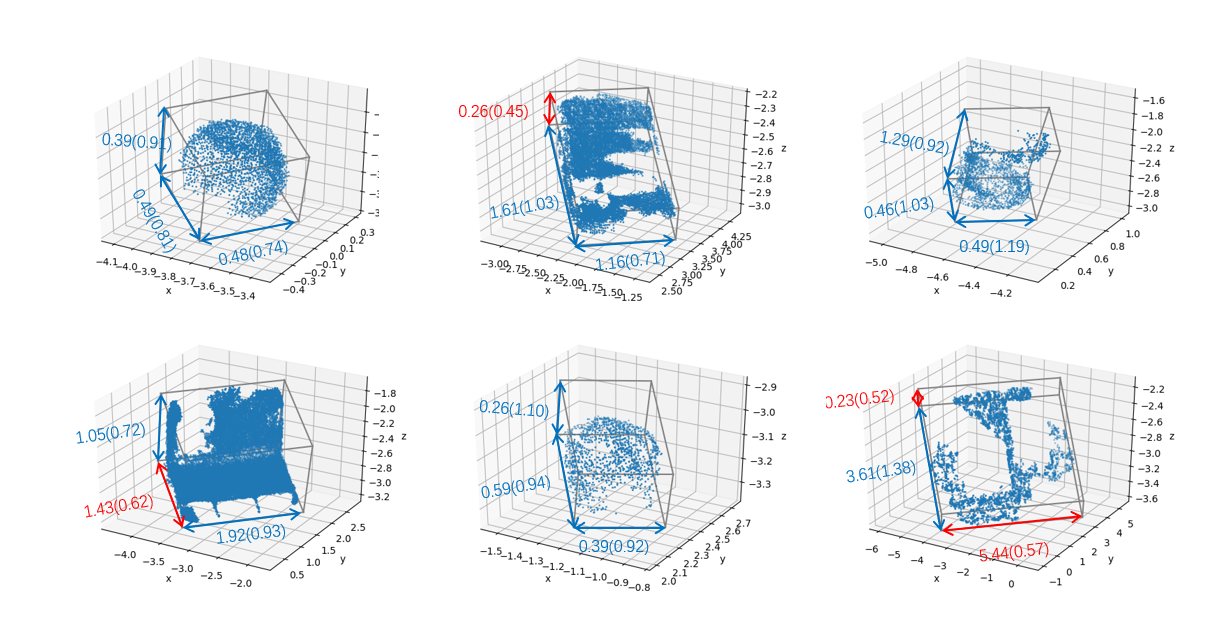}
    \caption{\hb{Top:} an example scene. \hb{Bottom:} the dimensions and confidence of different objects. The first row contains apple, notebook, and bottle. The second row contains laptop, mouse, and table. The data are annotated in the form of [dimension]([confidence]). Unreliable dimensions are marked red if we take 0.7 as the threshold.}
    \label{fig:dimension-example}
\end{figure}

We divide the 3D bounding box where the object locates into $8 \times 8 \times 8$ space grids and count the points in each grid. Let $\mathbb{G} = \{G_{(x, y, z)}\}_{x, y, z=1}^{x, y, z = 8}$ denote these grids and $N(G)$ denote the number of points $G$ contains where $x, y, z$ correspond to the direction of length, width and height, respectively. We define the global density of the point cloud as follows:
\begin{equation}
     \rho_g = \frac {\sum_{G \in M} N(G)} {|M|} \ ,\\  where\ M = \{G | G \in \mathbb{G}\ ,\  N(G) > 0 \} 
\end{equation}
and similarly define the density of point clouds on both sides of the length direction as follows: 
\begin{equation}
    \begin{aligned}
    \rho_{x_{head}} = \frac {\sum_{G \in M_{x_{head}}} N(G)} {|M_{x_{head}}|}\ , & \rho_{x_{tail}} = \frac {\sum_{G \in M_{x_{tail}}} N(G)} {|M_{x_{tail}}|}\ , \\
     where &\ M_{x_{head}} = \{G_{(x, y, z)}|G_{(x, y, z)} \in M, x = 1\}\ , \\
     &\ M_{x_{tail}} = \{G_{(x, y, z)}|G_{(x, y, z)} \in M, x = 8\}.
    \end{aligned}
\end{equation}
In this way we can determine the reliability of the beginning and end of the object in that direction, that is, the reliability of the dimension. The confidence (see Figure~\ref{fig:dimension-example} for \hb{an} example) of this dimension is defined as (similarly for other dimensions): 


\begin{equation}
    \eta_x = \frac 1 {\rho_g} \sqrt{\rho_{x_{head}} \cdot \rho_{x_{tail}}}. 
\end{equation}
When the confidence falls below a certain threshold, we assume that the results of this dimension are unreliable and accordingly do not use the distribution of this object in three dimensions in the scale optimization process, but instead degrade it to a two-dimensional distribution.

\begin{figure}[htbp]
    
    \centering
    \subcaptionbox{0 degree}
    {
    \includegraphics[width=0.3\columnwidth]{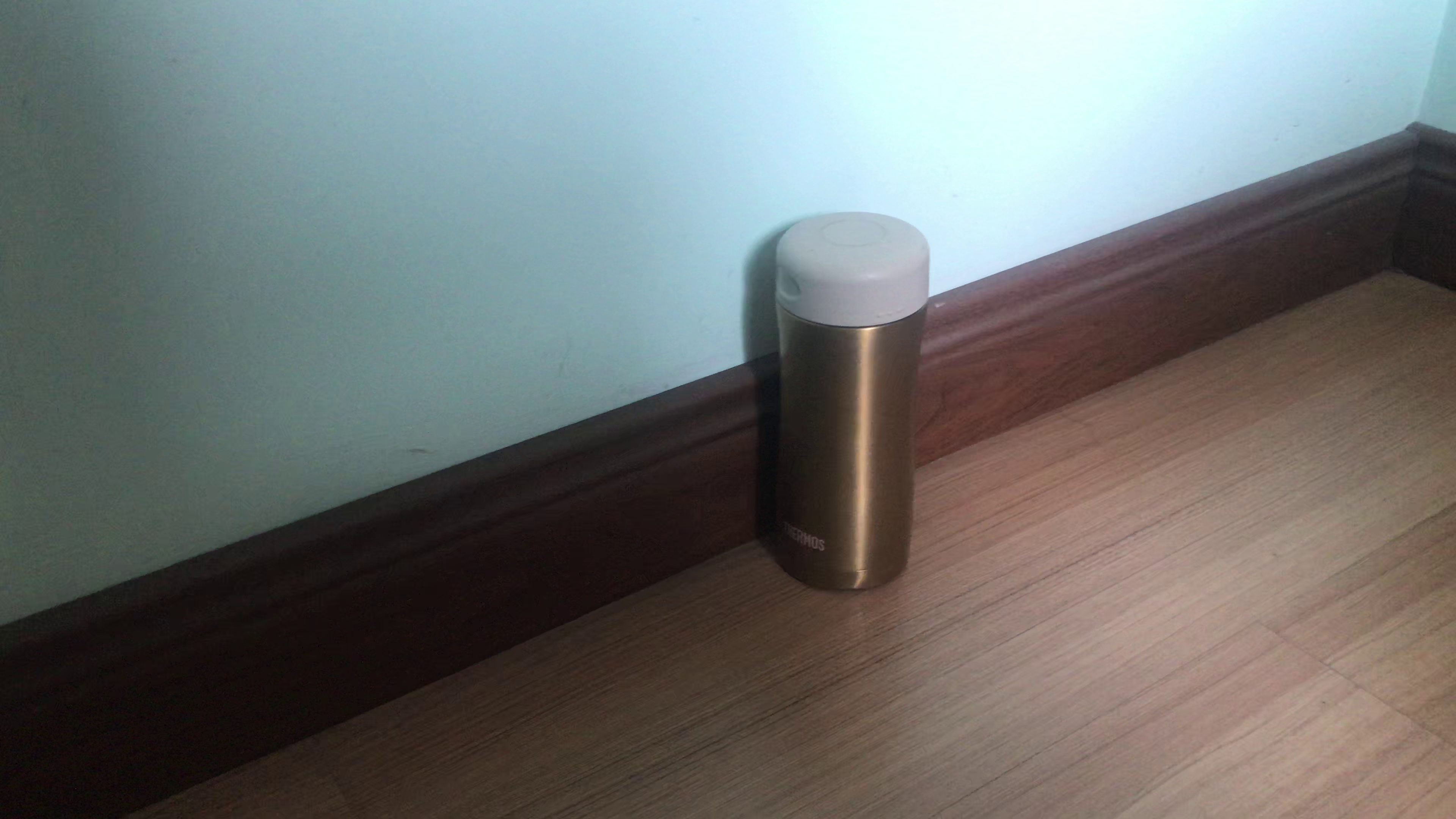}
    }
    \subcaptionbox{30 degrees}{
    \includegraphics[width=0.3\columnwidth]{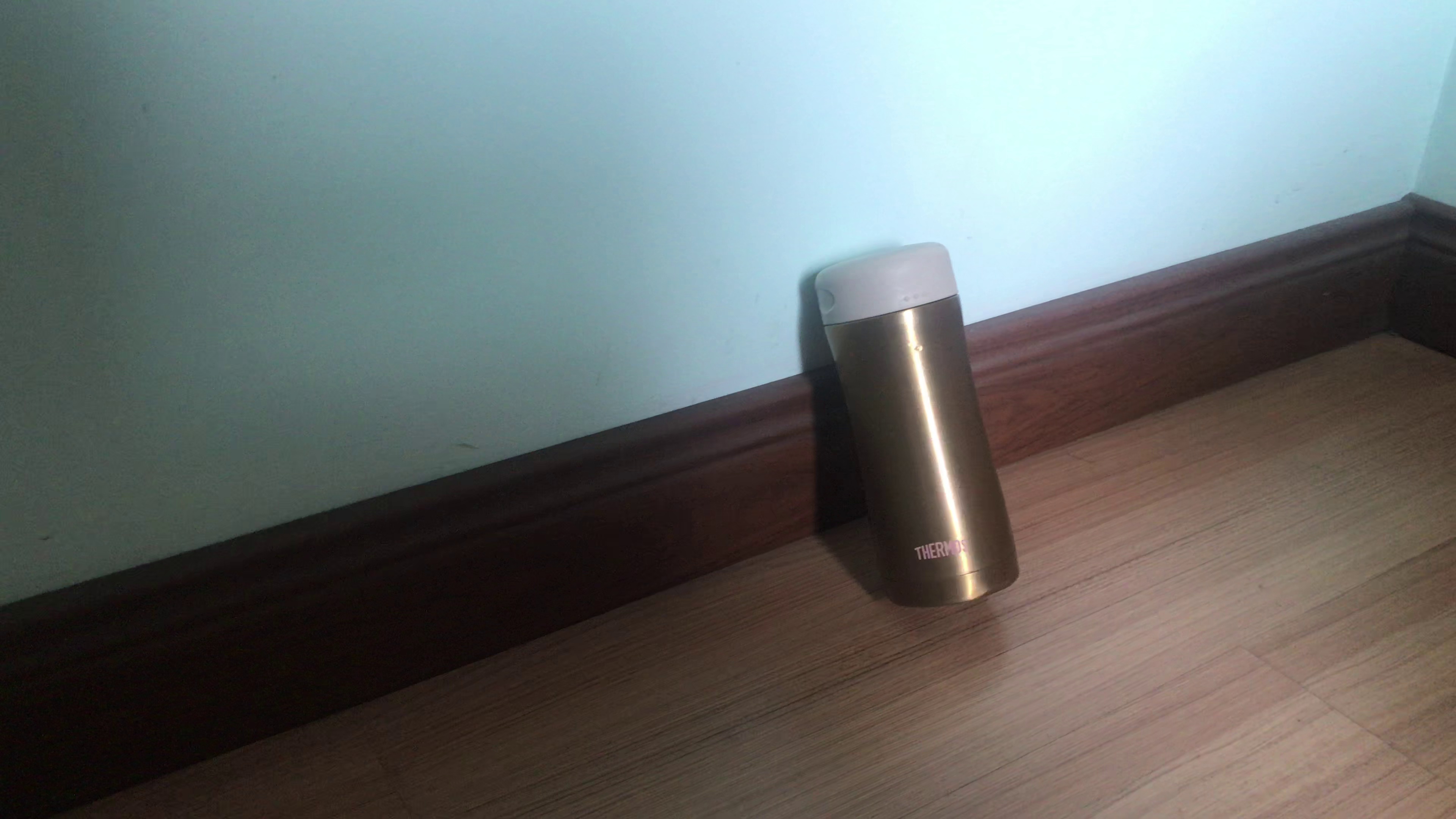}
    }
    \subcaptionbox{45 degrees}{
    \includegraphics[width=0.3\columnwidth]{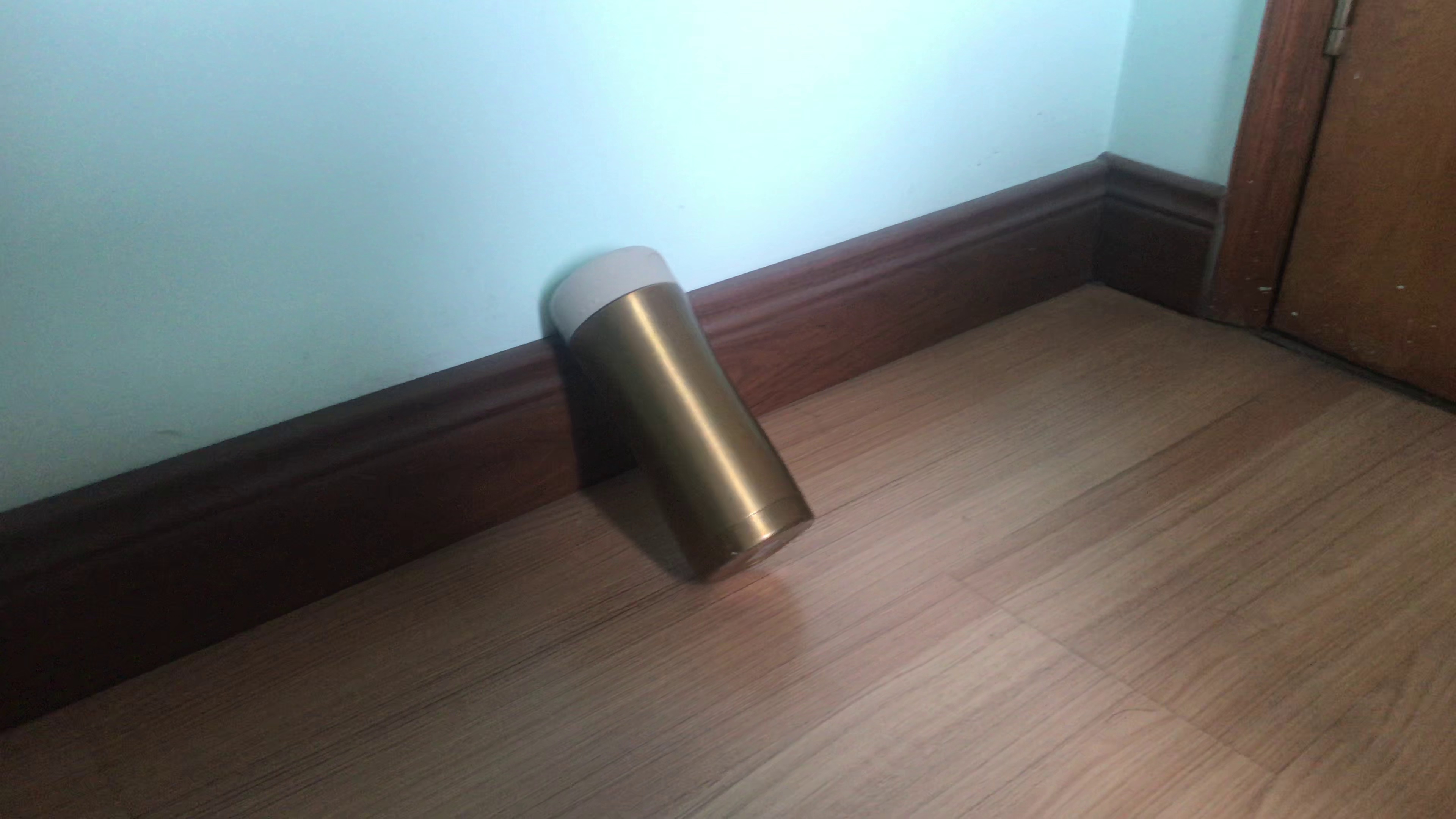}
    }
    
  \begin{threeparttable}
  \fontsize{6.5}{8}\selectfont
  \setlength{\tabcolsep}{6mm}{
    \begin{tabular}{cccc}
     \toprule
    Cases & $\eta_x$ & $\eta_y$ & $\eta_z$ \cr
    \midrule
     a & 0.912 & 1.077 & 0.926 \cr
    
     b & 0.885 & 1.001 & 0.543 \cr
    
     c & 0.593 & 1.045 & 0.291 \cr
    \bottomrule
    \end{tabular}
    }
    \end{threeparttable}
    \caption{Top: \hb{A} bottle is placed at three different angles relative to the vertical direction. The three cases are (a) 0 degree, (b) 30 degree\hb{s} and (c) 45 degree\hb{s,}  respectively. Bottom: confidence of three dimensions in the three cases.}
    \label{fig:three-cases}
\end{figure}

Usually, objects are placed upright on a horizontal plane. In such cases, our dimension extraction is accurate and efficient. when objects are tilted, the estimated height of these objects can be smaller and the length and width can be larger accordingly, and however  
the confidence on these dimensions will also go down (see Figure~\ref{fig:three-cases} for example), so that these dimensions tend not to be selected. Even there are few such dimensions extracted with errors, the accuracy of our algorithm will not decrease significantly, since we use all the objects in the scene to optimize the scaling factor.

\subsection{Scaling factor optimization}
\label{sec:ScaleOpt}

The reconstructed point cloud is similar with the real scene, and there is a proportional coefficient $s$ between them. For a distance estimated by the point cloud $l$, let $l^*$ denote the real distance and they have the following relationship:
\begin{equation}
    s\ =\ \frac {l^*} l. 
\end{equation}
We call $s$ as the scaling factor. The goal of scale optimization is to find the most likely scaling factor $s^*$ based on the statistical distribution of object sizes. Let $\{m_i\}_{i=1}^N$ denote the objects that appear in the scene, where $m_i$ has its dimensions as
\begin{equation}
    L_i = (w_i, l_i, h_i),\ i = 1, 2, \dots, N
\end{equation}

Let $\{\varphi_i(L)_{i=1}^N\}$ be the obtained size prior, each described by a GMM model. Assuming that the size of each object is independent of each other, with Bayesian Rule, we have

\begin{equation}
    \begin{aligned}
         & P(s|m_1, m_2, \dots, m_n)    \\
         & \propto P(s|m_1, m_2, \dots, m_{n-1})\ P(m_n|s) \\
         & \propto \prod_{i=1}^N\ P(m_i|s)                                           \\
         & \propto \prod_{i=1}^N\ \varphi_i(sL_i).                                    \\
    \end{aligned}
\end{equation}
Thus the best scaling factor $s^*$ is
\begin{equation}
    s^* = \argmax_{\forall s \in \mathbb{R^+}}\ \prod_{i=1}^N\ \varphi_i(sL_i).
\end{equation}
Since it is difficult to find mathematically precise optimal values for the above problems, we estimate the above probabilities on a series of discrete candidate values $s\in [s_{min}, s_{min}+ \Delta s, \dots, s_{max}]$, and select $\hat s$ with the maximum probability as the estimated scaling factor.  In the course of the experiment, it was found that usually $\prod_{i=1}^N\ \varphi_i(sL_i)$ is well convex near the maximum value, thus almost ensuring that $\abs{\hat s - s^*} < \Delta s$.

In the actual optimization, because some objects do not have high confidence in a certain dimension, or the certain dimension of some category of objects is insignificant or hard to collect, in such cases we do not use all three dimensions, but rather prescribe which dimensions of a certain type of objects to use as a priori distribution (e.g., length-width distribution for keyboard, height distribution for human, etc.), that is,  determining $\varphi_i$, and consider the calculated confidence to decide which dimensions of objects to use, that is, determining $L_i$.

\section{Results and discussion}

In this section, we conduct several experiments including comparisons
to the existing techniques to verify the effectiveness of our method. 
We also present virtual object insertion results as well as qualitative evaluation by a user study.
Finally, we perform additional experiments to show the necessity
of dimension extraction and the importance of richness of size priors to the accuracy of scaling factor optimization.

\begin{figure}
    \centering
    \includegraphics[width=\columnwidth]{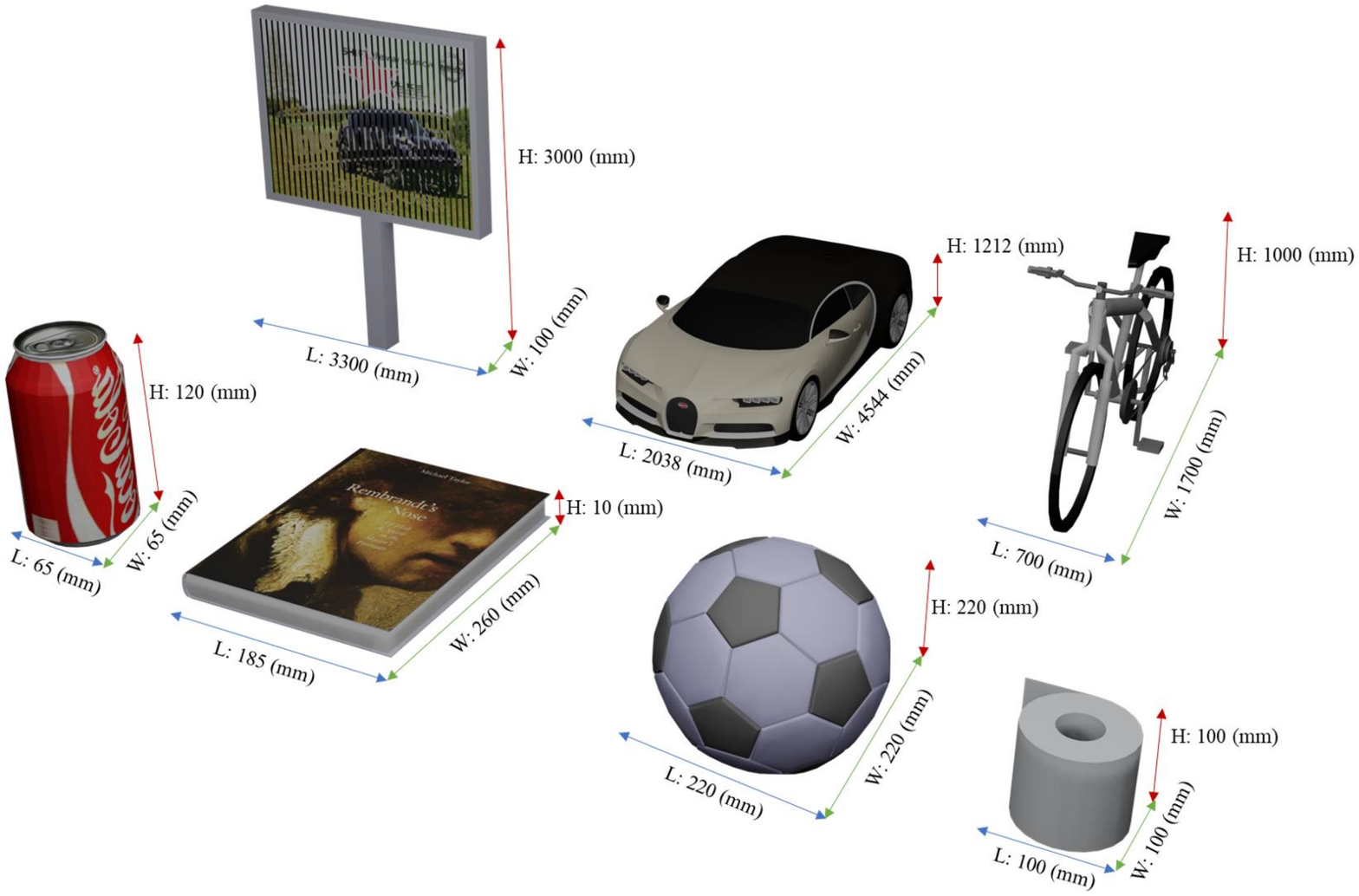}
    \caption{Inserted object\hb{s} with their physical sizes.}
    \label{fig:insertedobj}
\end{figure}

\begin{figure*}
    \centering
     \includegraphics[width=1.8in]{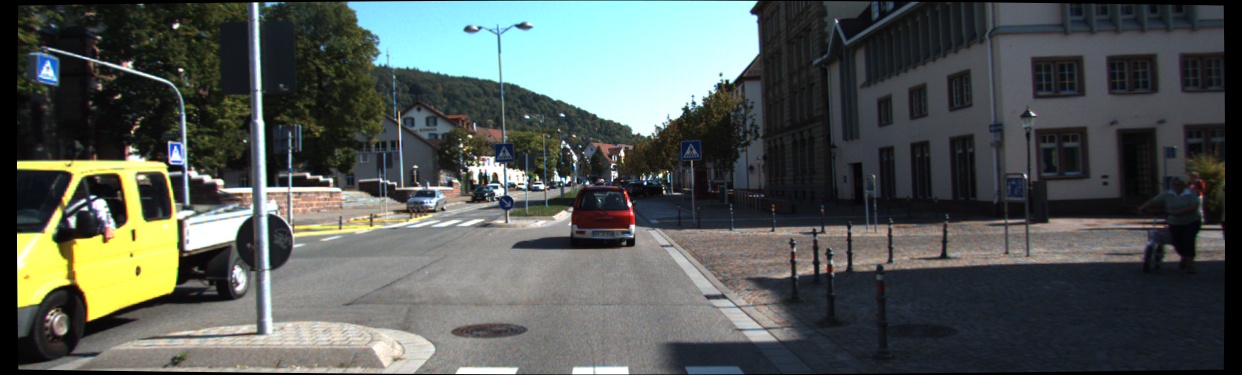}
     \includegraphics[width=1.8in]{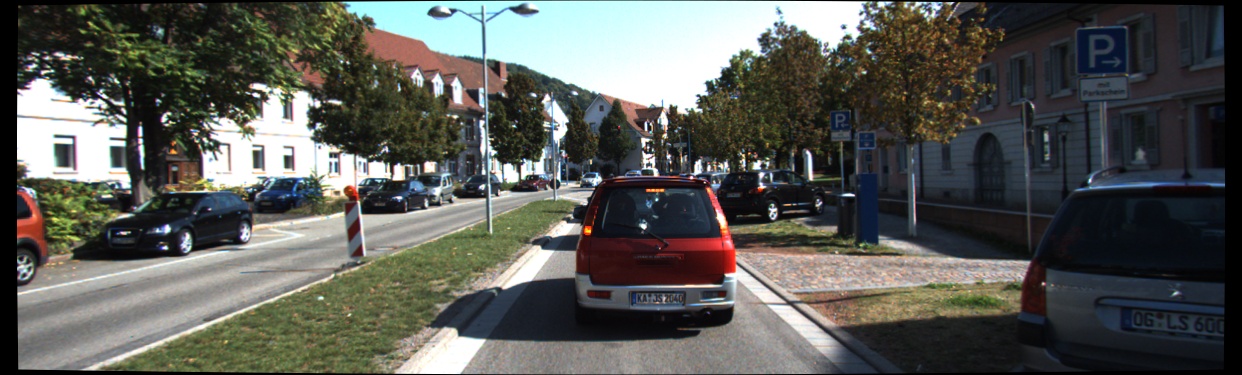}
     \includegraphics[width=1.8in]{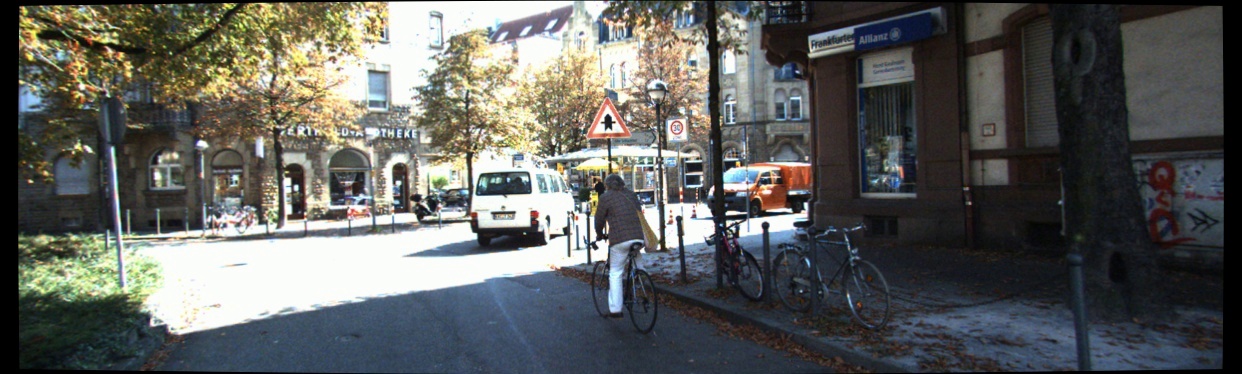}
     \\
     \includegraphics[width=1.8in]{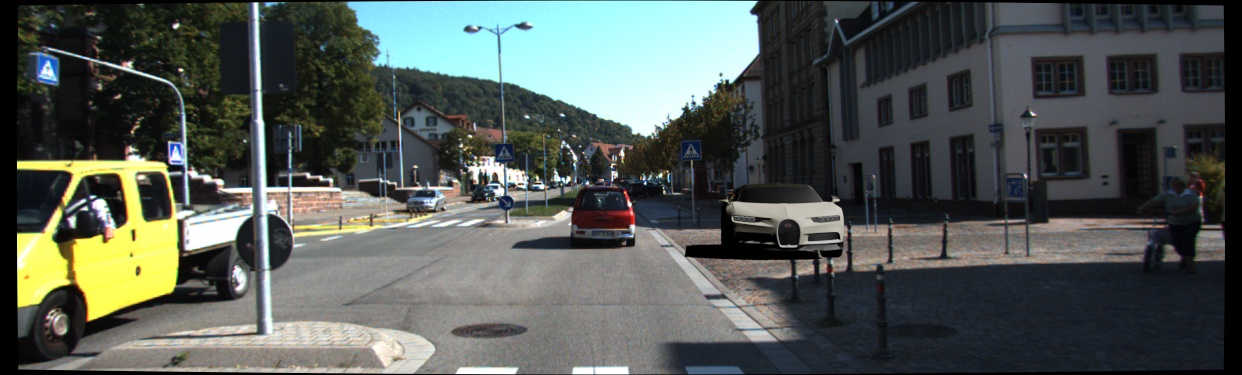}
     \includegraphics[width=1.8in]{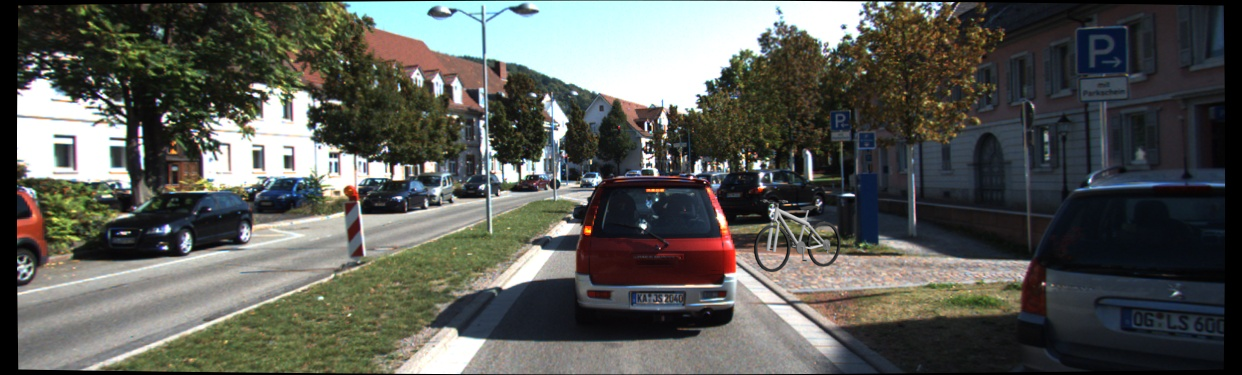}
     \includegraphics[width=1.8in]{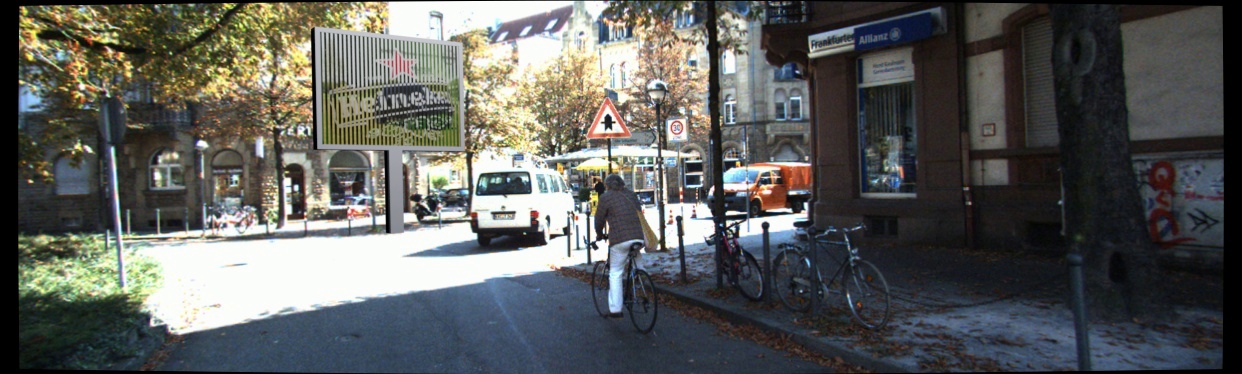}
     \\
     
      \includegraphics[width=1.8in]{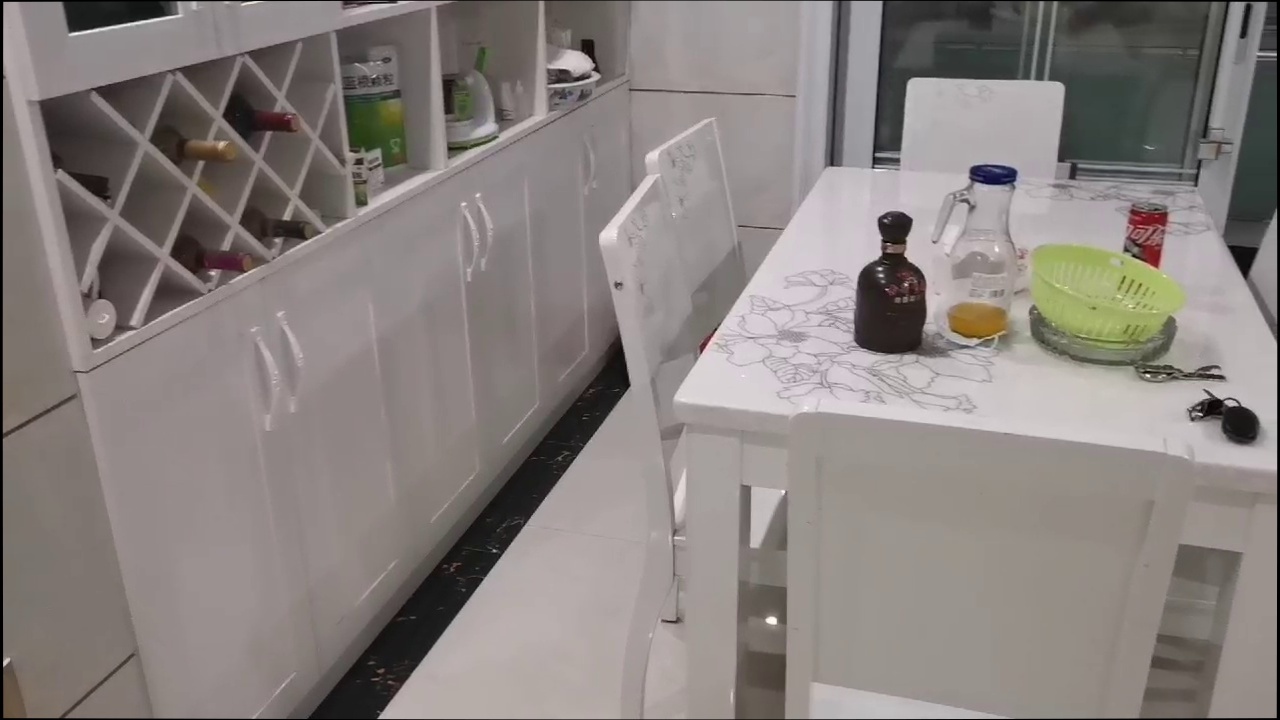}
      \includegraphics[width=1.8in]{pictures/insertion/8.jpg}
      \includegraphics[width=1.8in]{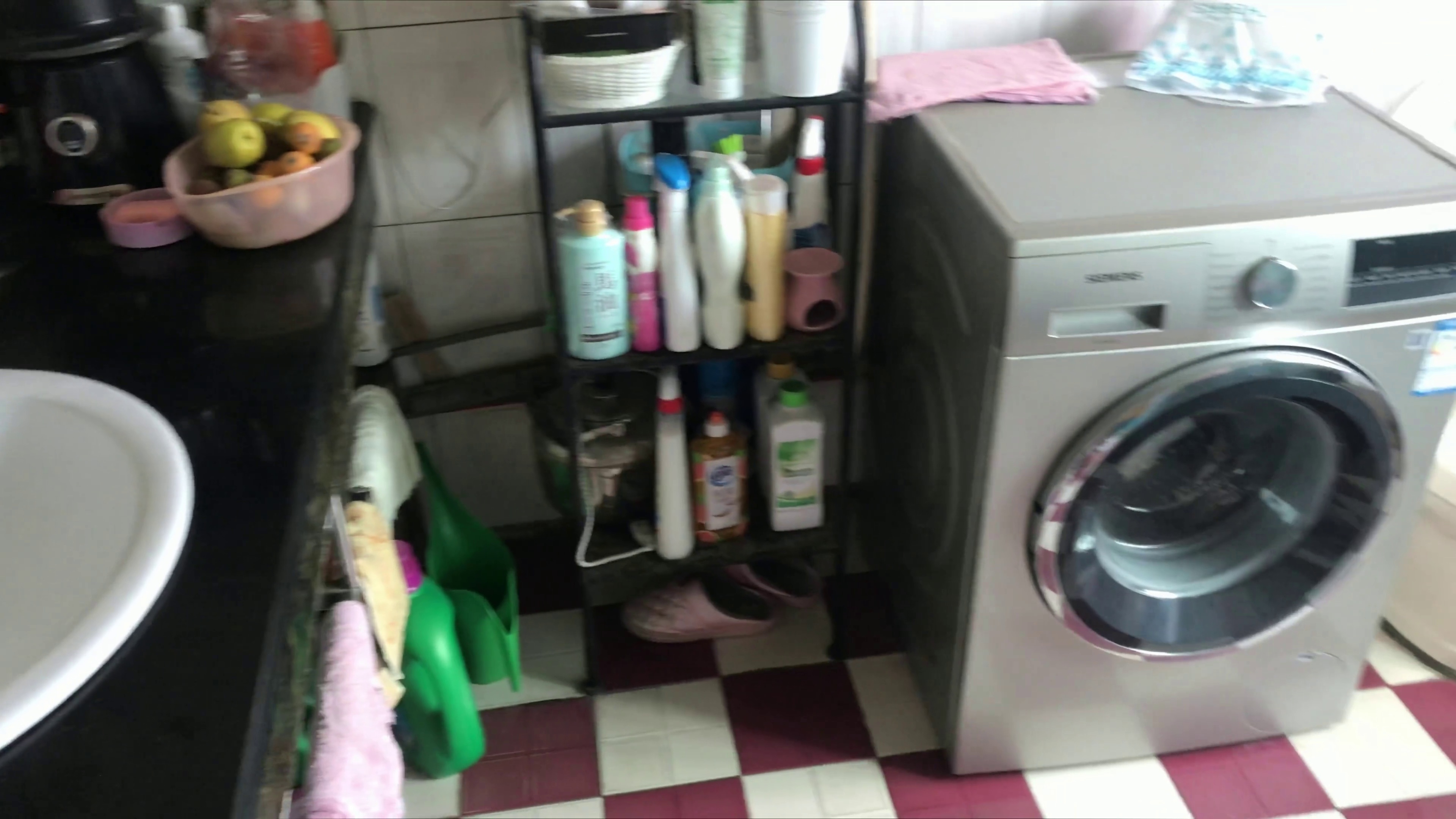}
      \\
     
      \includegraphics[width=1.8in]{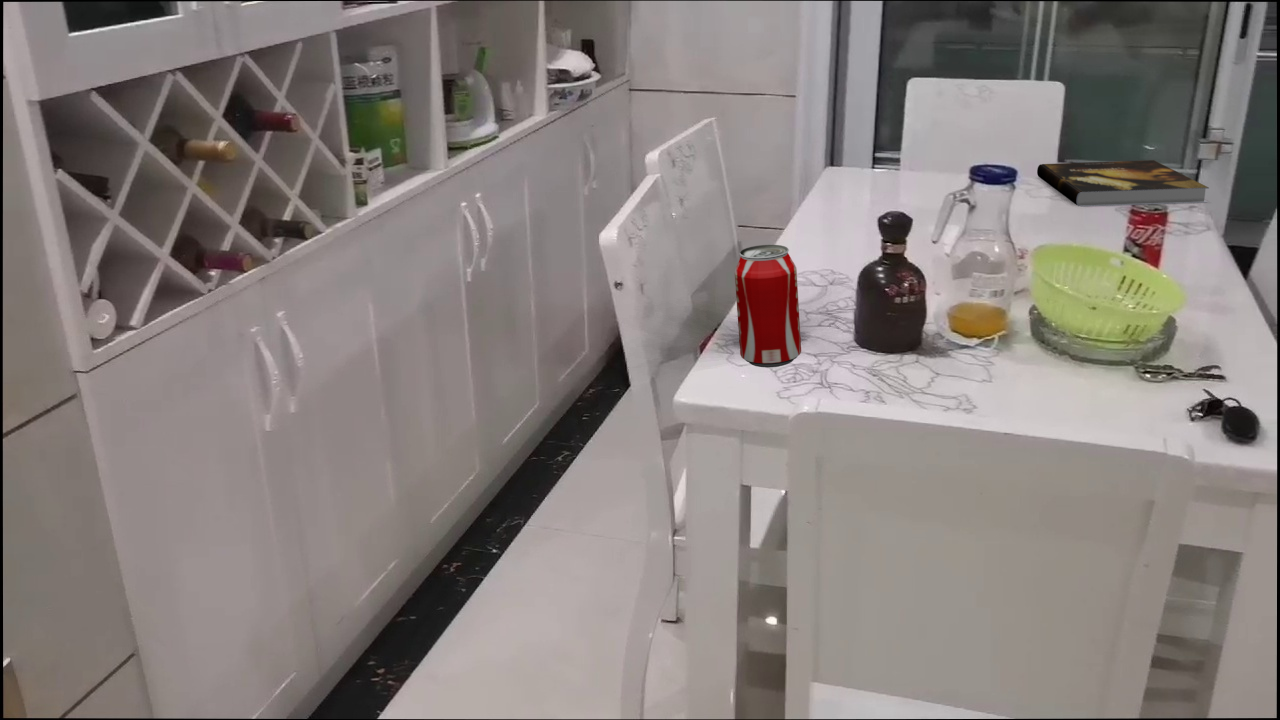}
      \includegraphics[width=1.8in]{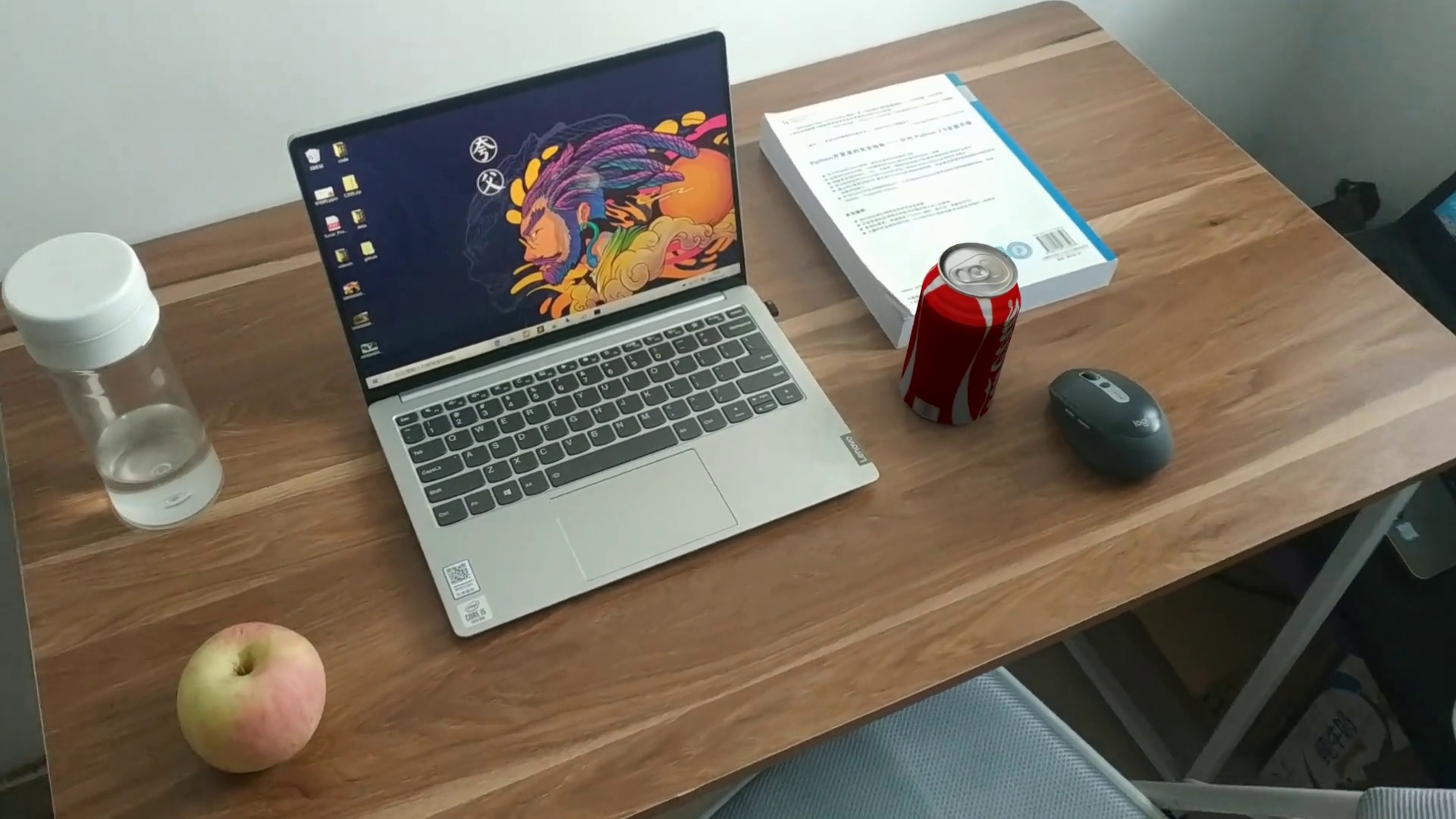}
      \includegraphics[width=1.8in]{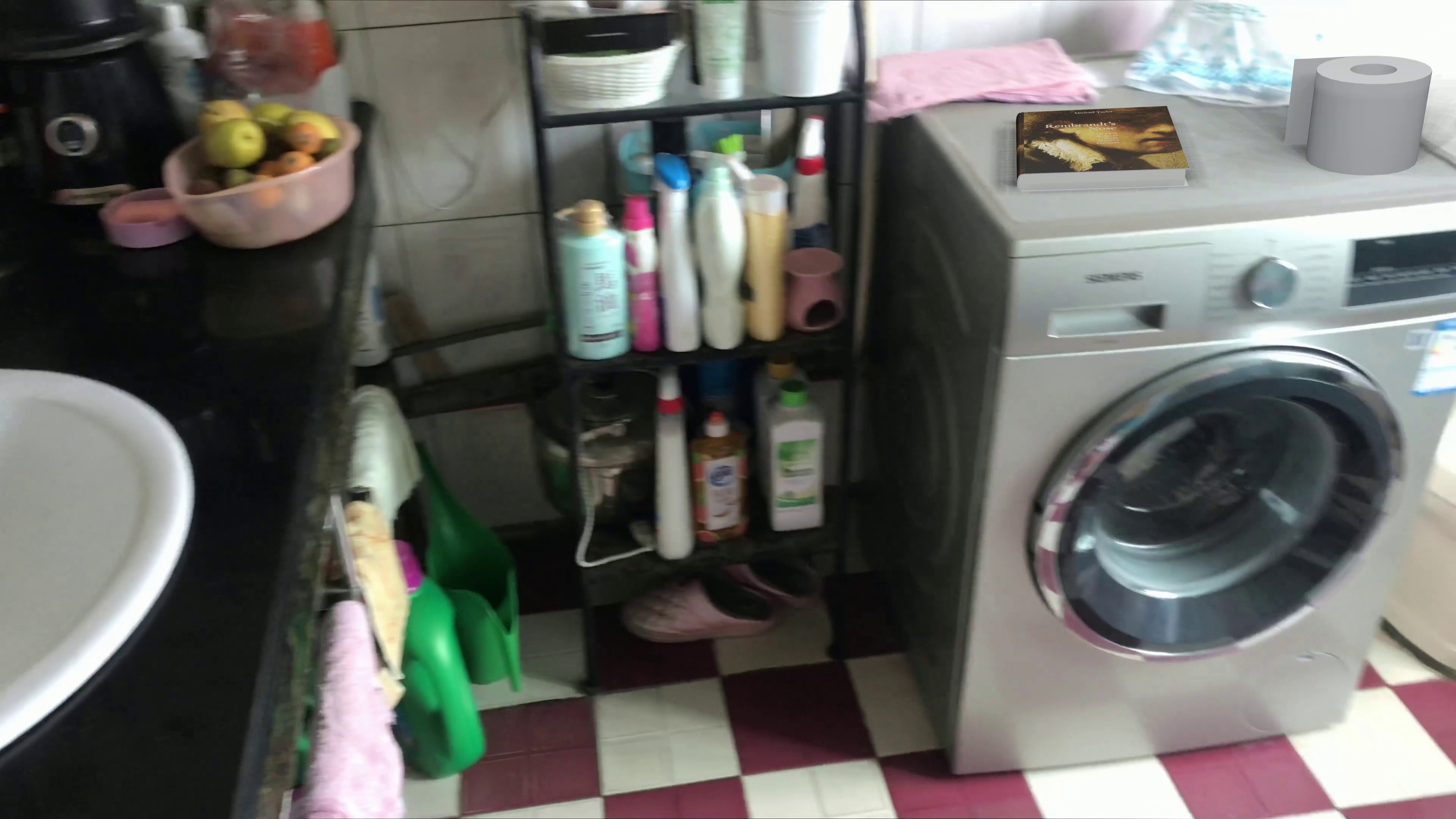}
    
    \caption{Synthesis results of insertion of virtual objects in Fig.\ref{fig:insertedobj}. Rows 1 and 3 are input videos and Rows 2 and 4 are output videos.}
    \label{fig:result}
\end{figure*}

\begin{figure}
    \centering
    \includegraphics[width=.8\columnwidth]{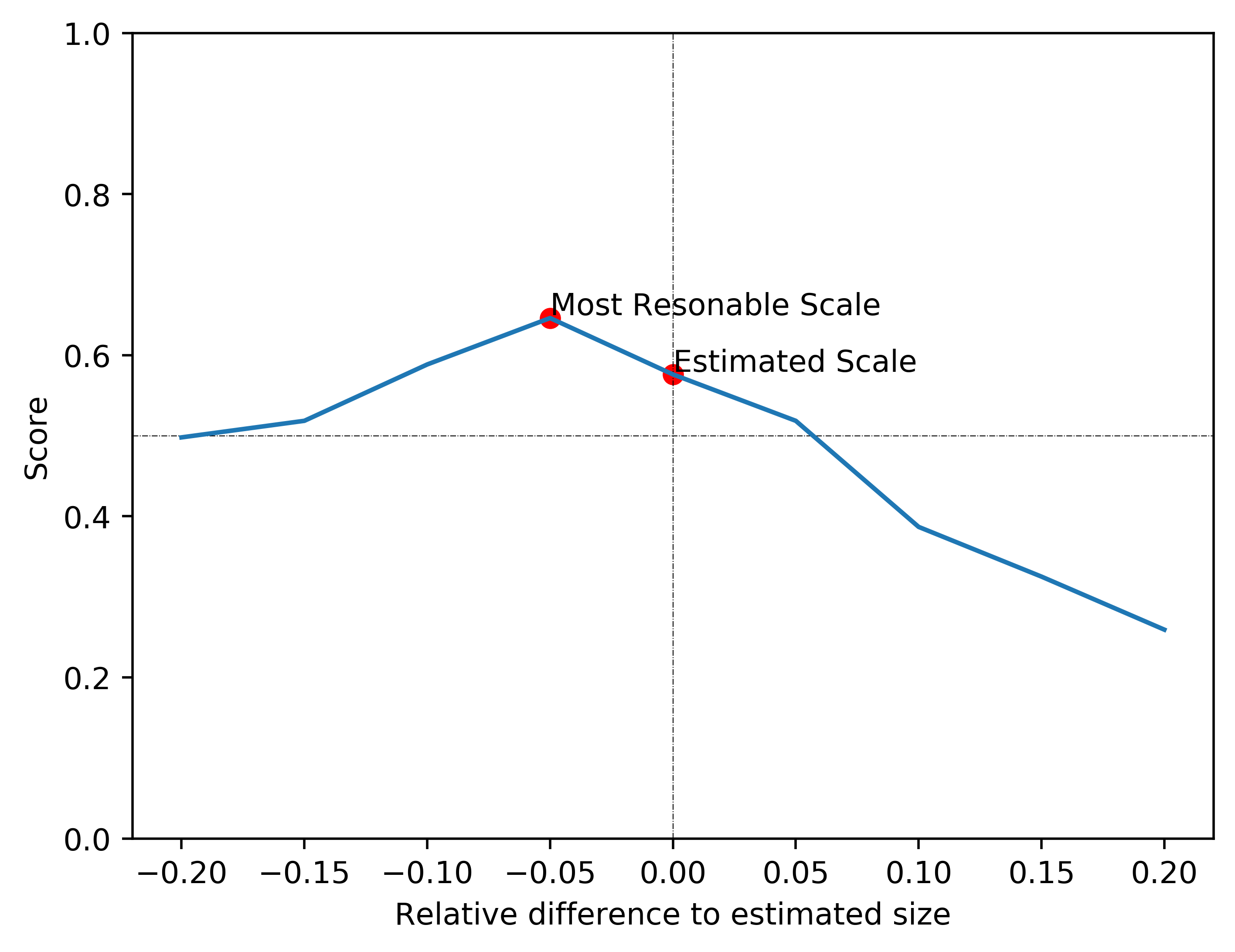}
    \caption{The results of user study.}
    \label{fig:res-userstudy}
\end{figure}

\subsection{Comparison on scene size estimation}
\label{sec:compare}
We choose \cite{sucar:2017:probabilistic} as our baseline, since the goal of this work is similar to ours, which is to calculate the global scale of the scene. 
We compare the performance of scale estimation by these two approaches on two datasets: the Kitti raw data\cite{Geiger2013IJRR} as an outdoor dataset and our captured indoor video dataset.

\textbf{Datasets.} To demonstrate the robustness and superiority of our method, we use both an indoor dataset and an outdoor dataset for comparison. For the indoor dataset, we captured 5 indoor videos including sitting room, kitchen, toilet, bedroom and study and measured the sizes of objects in these videos as a validation set. For the outdoor dataset, We choose 18 video sequences from the Kitti raw Data supported by Geiger et al. \cite{Geiger2013IJRR} according to the segmentation results. This outdoor dataset contains processed color stereo sequences and 3D Velodyne point clouds, with the known camera parameters and 3D object tracklet labels.

\textbf{Comparison.} For fair comparison, we use Mask RCNN as the 2D object detector and OpenSfM as the reconstruction module for our approach and \cite{sucar:2017:probabilistic}. We calculate the accuracy of the global scale by using the relative error of the estimated object sizes and the physical object sizes for validation. In our experiments, we use millimeter as the unit. Since the method in \cite{sucar:2017:probabilistic} can only estimate the heights of objects, we use the mean of the relative error between the estimated object height and the ground-truth height over all objects as the error metric. 
 The comparison results in Table \ref{tab:performance_comparison} show that the relative error of our algorithm is nearly 10\% lower than the baseline, reflecting the robustness and superiority of our method.

\textbf{Processing timings.} We tested the performance of our method on a PC with i7-6850k CPU and Nvidia 1080 Ti, 32G RAM. In our experiments, video frames range from 100 to 400, and the processing time of OpenSfM is relative to the square of the number of frames. For example, if there are about 100 video frames, OpenSfM takes about 36 mins. The processing time of other modules is relatively fixed. Specifically, instance segmentation and point cloud segmentation take about 0.3s and 0.2s per frame, respectively. Dimension extraction takes about 0.02s, and the optimization step takes about 0.004s for an entire video. Therefore, a real-time vSLAM and a lightweight instance segmentation implementation would greatly improve the overall performance.

\begin{table}[tp]
  \centering
  \fontsize{6.5}{8}\selectfont
  \begin{threeparttable}
  \caption{Performance comparison on video datasets.}
  \label{tab:performance_comparison}
    \begin{tabular}{cccccc}
    \toprule
    Method&Amount
    &\multicolumn{2}{c}{Sucar et al.}&\multicolumn{2}{c}{Ours}\cr
    \cmidrule(lr){3-4} \cmidrule(lr){5-6}
    &&Err&Std&Err&Std\cr
    \midrule
    Indoor&5&0.168&0.049&{\bf 0.072}&0.025\cr
    Outdoor&18&0.163&0.099&{\bf 0.074}&0.034\cr
    \bottomrule
    \end{tabular}
    \end{threeparttable}
\end{table}

\subsection{Virtual object insertion results}
We present 10 synthesis virtual object insertion results for indoor and outdoor scenes with automatic determination of the sizes of the inserted objects, shown in Fig. \ref{fig:teaser} and  \ref{fig:result}. The corresponding videos can be found in the supplementary materials. Note that currently we manually choose the position in the scenes.

We conduct a user study to evaluate the synthesis results qualitatively. We sample 30 frames from the above synthesis results (3 for each) containing the inserted virtual objects from near, medium and far distant for outdoor videos and from different views for indoor videos. Fig. \ref{fig:result} shows a few representative frames.

For each frame, we also generate the synthesis results for the same virtual objects at the 
same positions with only the difference in size with the interval [-30\%,-20\%,-15\%,-10\%,-5\%,0\%,5\%,10\%,15\%,20\%,30\%] relative to our estimated sizes.
Then 27 subjects were invited to judge whether or not the inserted objects in the synthesis results are reasonable in size by answering Yes or No. All the subjects were university students above the age of 20.

Figure \ref{fig:res-userstudy} illustrates the average scores of all scenes from all participants. The horizontal axis corresponds to the the difference in size relative to our estimated sizes, and the vertical axis is the averaged score (1 for YES) for all participants. 

The peak of the curve is 0.65, showing that human size judgments have significant variation, and can be influenced by context and familiarity\cite{Predebon1992The}.
Despite the variation, there are still some useful conclusions from the curve.
First, our score is above 0.5 showing that more than half participants agree with our results.
And the peak is on the -0.05 of X-axis, which means that the ground truth of size appears at 0.95 times our estimated sizes, showing that the scale error of our method is around 5\%.

\subsection{Discussion}

\textbf{With/without dimension extraction.} There is a simple strategy by sending all the three dimensions of segmented objects in point clouds to scaling factor optimization in order to provide more hints on sizes.
This 
is helpful apparently only for relatively good 3D reconstruction results from carefully captured videos, for example, with the camera moving around 
a center of the scene.
However, most of the captured videos involve more complicated camera movement, causing only partial geometry recovered from the videos.
We did an experiment on the same scenes in scaling factor optimization with the extracted dimensions or directly with the dimensions of bounding boxes. The results show that the relative error of the scaling factor increases from 7\% to 30\%. 

\begin{figure}
    \centering
    \includegraphics[width=.9\columnwidth]{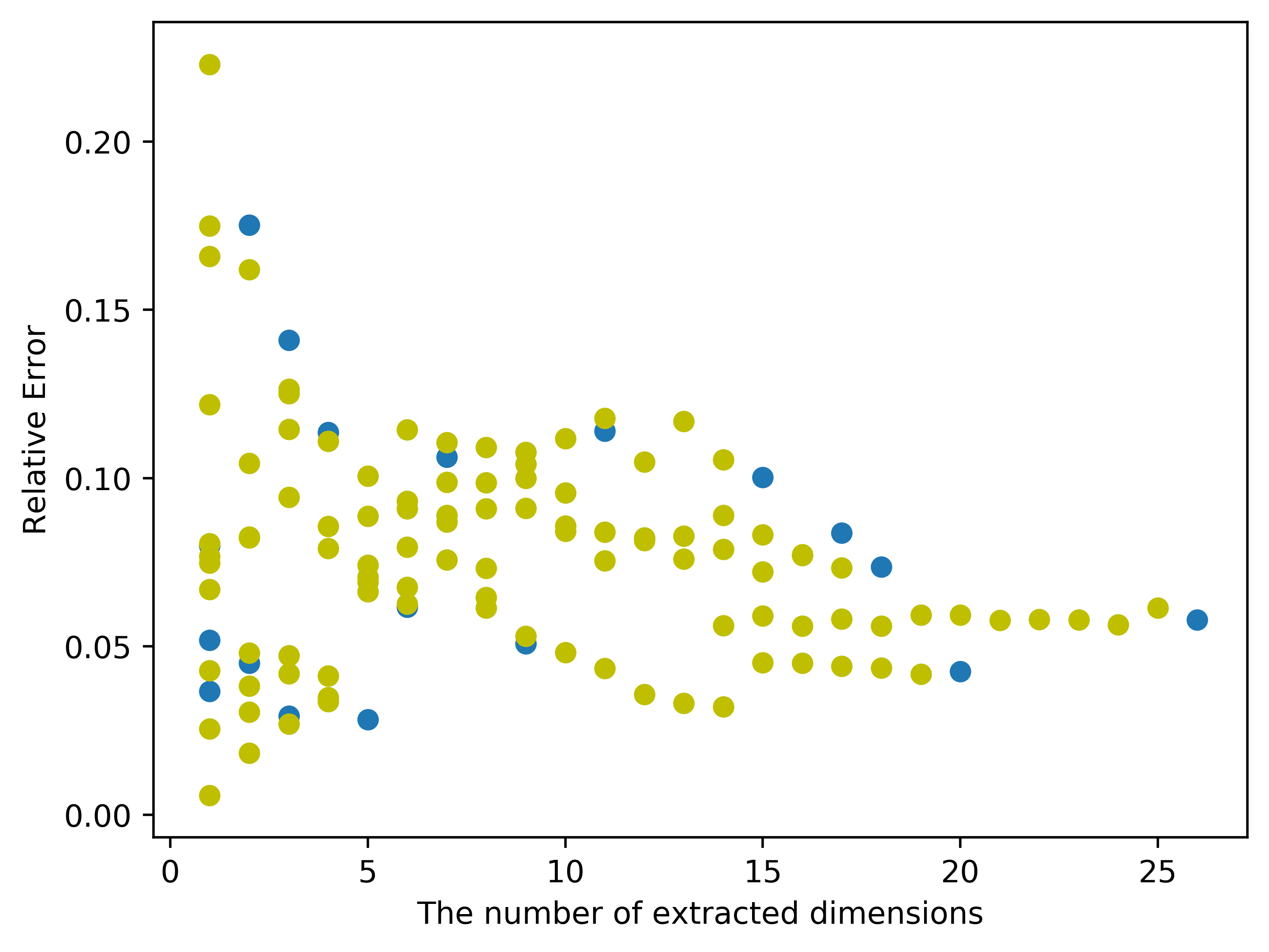}
    \caption{The relation between extracted dimensions and scale accuracy.}
    \label{fig:relation-dimandacc}
\end{figure}

 \textbf{Limitations.} Our method estimates actual object sizes in the process of 3D reconstruction, and thus would fail when OpenSfM fails, e.g., due to static views or highly dynamic scenes. We assume that the intrinsic parameters of the camera are fixed and the roll of the camera is zero,  which are common cases in real life, so that scene object sizes should strictly conform to the same global scale. Our method can cope with scenes with some objects that are not upright, but  will fail if a  scene is totally in a mess.

\textbf{Relation between scale accuracy and the number of extracted dimensions.}
We show some experiment results to illustrate how the estimated accuracy of scale is related to the number of extracted dimensions of objects incorporated in scale optimization. 
Fig. \ref{fig:relation-dimandacc} shows the relation between extracted dimensions and scale accuracy of 23  scenes. In this experiment, we gradually decrease the number to 3 of extracted dimensions for simulation and illustrated as scatter plot. The figure shows the rapidly descending trends of scale accuracy along with the number of extracted dimensions.

\begin{figure}
    \centering
    \includegraphics[width=\columnwidth]{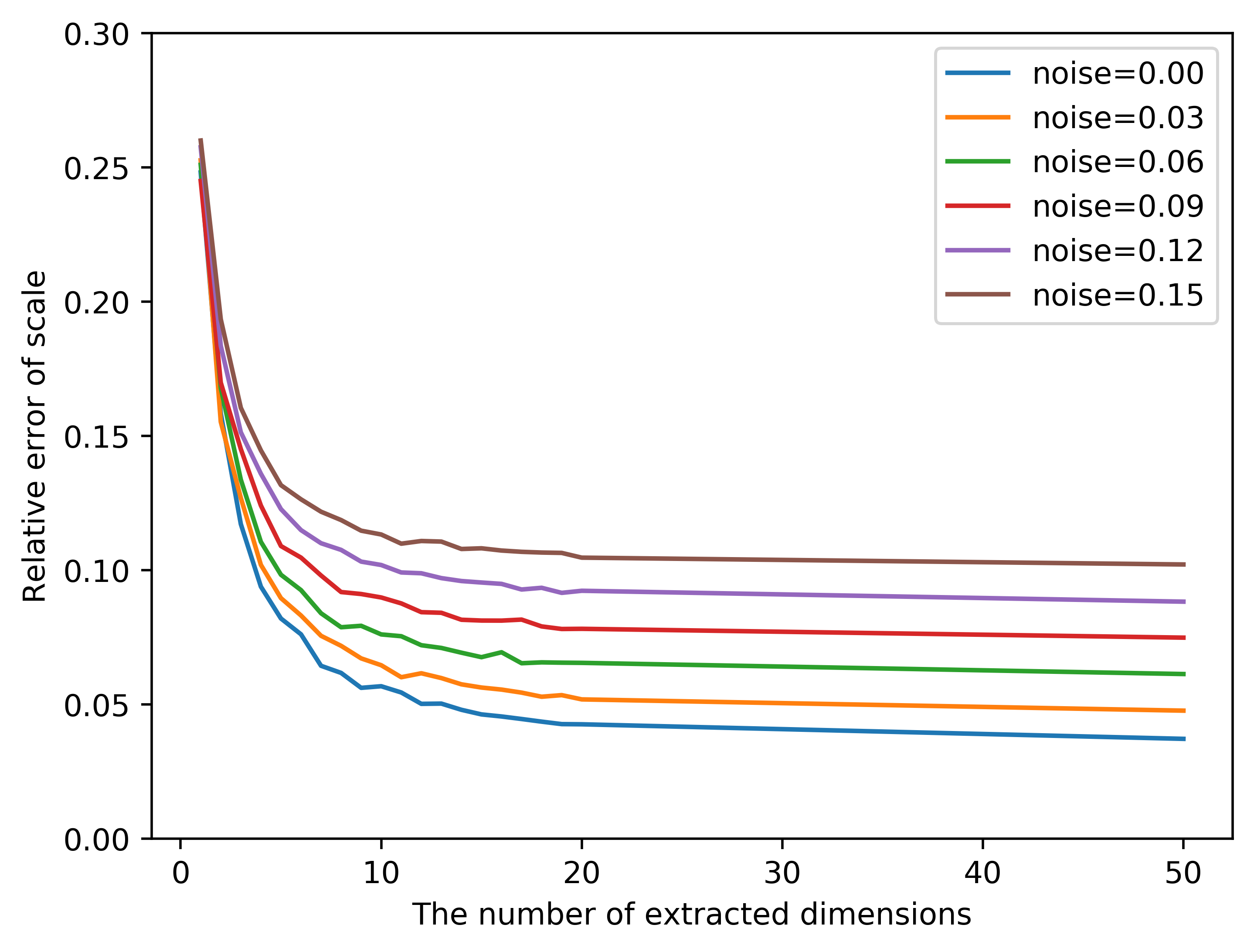}
    \caption{The scale accuracy in simulation data.}
    \label{fig:relation-simulation}
\end{figure}

We did a simulated experiment on the Structure3D dataset \cite{zheng2019structured3d}.
This dataset contains 3,500 house designs and over 343 thousand instances in 24 categories.
The size distributions of all 24 categories are calculated by instance size statistics.
We select different numbers $N$ of objects randomly from each house, \hb{then} disturb the sizes of selected objects in no more than a relative error $R$, and 
\hb{finally} estimate the average of scale of each house.
Fig. \ref{fig:relation-simulation} is drawn with $N$ in range of $[1-10,20,50]$ and $R$ in range of $[0,3\%,6\%,9\%,12\%,15\%]$.
The results provide us several interesting observations, which can guide the choice according to the performance of semantic segmentation and 3D reconstruction: 
\begin{itemize}
    \item The scale error decreased rapidly when the number of recognized objects is below 10.
    \item The accuracy of bounding box of recovered objects in scenes is a critical factor to scale estimation, so that the extraction of plausible dimensions is quite important to the incomplete and inaccurate recovered geometry.
    \item In the case of the same number of recognized objects, our scale accuracy in inaccurate geometry can compete the simulated results depending on the precise size distribution, showing that a fine-grained classification method is preferred to indicate a more precise size distribution of objects.
\end{itemize}

\section{Conclusion and future work}
With the aid of the object size distribution in Metric-Tree, we are able to make relatively accurate scale estimates of the scenes in monocular videos without other size inputs, resulting in the scale-aware object insertion. And the experiments on real scenes show that our method is a significant improvement on the scale estimation problem relative to similar previous work. Besides, the user study indicates that such virtual object insertion results are consistent with users' perceptions of scales.

Metric-Tree with a large physical size information as priors can be applied to more and more visual fields, including scale drift correction in automatic driving drift, layout scheme optimization in 3D scene synthesis, object pose optimization for object pose estimation. The collected images are helpful to fine grained classification, and conversely, the scale estimation also needs a fine grained classification method to indicating more precise size distribution of the objects.

\acknowledgments{
The authors would like to thank all reviewers for their thoughtful
comments. This work was supported by the National Key Technology
R\&D Program (Project Number 2017YFB1002604), the National Natural Science Foundation of China (Project Numbers 61521002, 61772298), Research Grant of Beijing Higher Institution
Engineering Research Center, and Tsinghua–Tencent Joint Laboratory for Internet Innovation Technology.}

\bibliographystyle{abbrv-doi}

\bibliography{template}
\end{document}